\documentclass[journal]{IEEEtran}
\usepackage{cite}
\usepackage{graphicx}
\usepackage{amsmath}
\usepackage{amssymb}
\usepackage{subcaption}
\usepackage{booktabs}
\usepackage{float}
\usepackage{multirow}
\usepackage[ruled,vlined]{algorithm2e}
\usepackage{titlesec}
\usepackage[font=footnotesize]{caption}
\usepackage{array}
\usepackage{tabularx} 
\usepackage{array}

\ifCLASSINFOpdf
\else
\fi
\begin{document}

\title{Free Energy-Inspired Cognitive Risk Integration for AV Navigation in Pedestrian-Rich Environments}
%
%
%

\author{Meiting Dang,
        Yanping Wu,
        Yafei Wang,~\IEEEmembership{Member,~IEEE},
        Dezong Zhao,~\IEEEmembership{Senior Member,~IEEE},
        David Flynn,~\IEEEmembership{Senior Member,~IEEE},
        and Chongfeng Wei*,~\IEEEmembership{Senior Member,~IEEE}

\thanks{*Chongfeng Wei is the corresponding author.}    
\thanks{Meiting Dang, Yanping Wu, Dezong Zhao, David Flynn and Chongfeng Wei are with James Watt School of Engineering, University of Glasgow, Glasgow, G12 8QQ, United Kingdom (email: m.dang.1@research.gla.ac.uk, 3066431W@student.gla.ac.uk, dezong.zhao@glasgow.ac.uk, David.Flynn@glasgow.ac.uk, chongfeng.wei@glasgow.ac.uk)}
\thanks{Yafei Wang is with the School of Mechanical Engineering, Shanghai Jiao
Tong University, Shanghai 200240, China (email: wyfjlu@sjtu.edu.cn)}}


%

\maketitle

\begin{abstract}
Recent advances in autonomous vehicle (AV) behavior planning have shown impressive social interaction capabilities when interacting with other road users. However, achieving human-like prediction and decision-making in interactions with vulnerable road users remains a key challenge in complex multi-agent interactive environments. Existing research focuses primarily on crowd navigation for small mobile robots, which cannot be directly applied to AVs due to inherent differences in their decision-making strategies and dynamic boundaries. Moreover, pedestrians in these multi-agent simulations follow fixed behavior patterns that cannot dynamically respond to AV actions. To overcome these limitations, this paper proposes a novel framework for modeling interactions between the AV and multiple pedestrians. In this framework, a cognitive process modeling approach inspired by the Free Energy Principle is integrated into both the AV and pedestrian models to simulate more realistic interaction dynamics. Specifically, the proposed pedestrian Cognitive-Risk Social Force Model adjusts goal-directed and repulsive forces using a fused measure of cognitive uncertainty and physical risk to produce human-like trajectories. Meanwhile, the AV leverages this fused risk to construct a dynamic, risk-aware adjacency matrix for a Graph Convolutional Network within a Soft Actor-Critic architecture, allowing it to make more reasonable and informed decisions. Simulation results indicate that our proposed framework effectively improves safety, efficiency, and smoothness of AV navigation compared to the state-of-the-art method.
\end{abstract}


%
\IEEEpeerreviewmaketitle

\section{Introduction}
\IEEEPARstart{I}{n} recent years, rapid advancements in autonomous driving technology have enabled autonomous vehicles (AV) to expand beyond simple, structured highway environments and to be increasingly deployed in more complex urban environments \cite{r22}. They are expected to become a key component of future urban transportation systems \cite{r21}. Unlike structured roads with clear traffic rules and physical lane separations, shared spaces in cities such as squares, campuses, and residential areas usually lack right-of-way regulations and explicit physical boundaries \cite{r23} between vehicles and pedestrians. In such environments, AVs must frequently interact with multiple pedestrians \cite{r24} and safely complete the navigation task despite ambiguities in road priority. The high uncertainty and dynamic nature of human behavior \cite{r25}, especially when multiple pedestrians are moving at the same time, significantly complicate the AV’s decision-making process. To ensure both safety and efficiency, AVs must make real-time decisions and continuously adapt their strategies in response to surrounding pedestrian behaviors. This poses a major challenge to existing AV decision-making systems. Hence, this work focuses on the decision-making problem of AVs interacting with multiple pedestrians in shared spaces, aiming to develop a safe and efficient decision-making model capable of handling multi-agent interactions and complex environmental uncertainties. \par
While numerous studies have explored crowd navigation for small-scale mobile robots \cite{r26, r27}, relatively few have focused on AVs operating in similar pedestrian-rich environments. Although AVs represent a specialized class of robotic systems, their navigation differs from that of smaller robots. Unlike service robots, AVs have larger physical footprints, operate at higher speeds, and are subject to more complex kinematic and safety constraints \cite{r28}. These distinctions make it impractical to directly apply robot navigation strategies to AVs. As urban environments increasingly require AVs to coexist with pedestrian crowds, there is a growing need for decision-making frameworks that support safe and efficient navigation in such complex settings.\par
Accurate pedestrian modeling plays a critical role in enabling realistic and adaptive AV decision-making in interactive environments. Many studies have adopted overly simplified pedestrian models with fixed behavior patterns, such as constant velocity assumptions \cite{r1} or pre-recorded real-world trajectories \cite{r2, r30}. These approaches commonly treat pedestrians as passive agents who are unable to dynamically respond to the AV’s behavior during simulation. Consequently, they fail to capture the bidirectional adjustment mechanisms essential for realistic interaction modeling. This limitation weakens the ability to model dynamic adaptive behaviors, which are essential for capturing the nature of human-vehicle interactions in shared environments. \par
\begin{figure*}[h]
\centering
\includegraphics[width=0.98\textwidth]{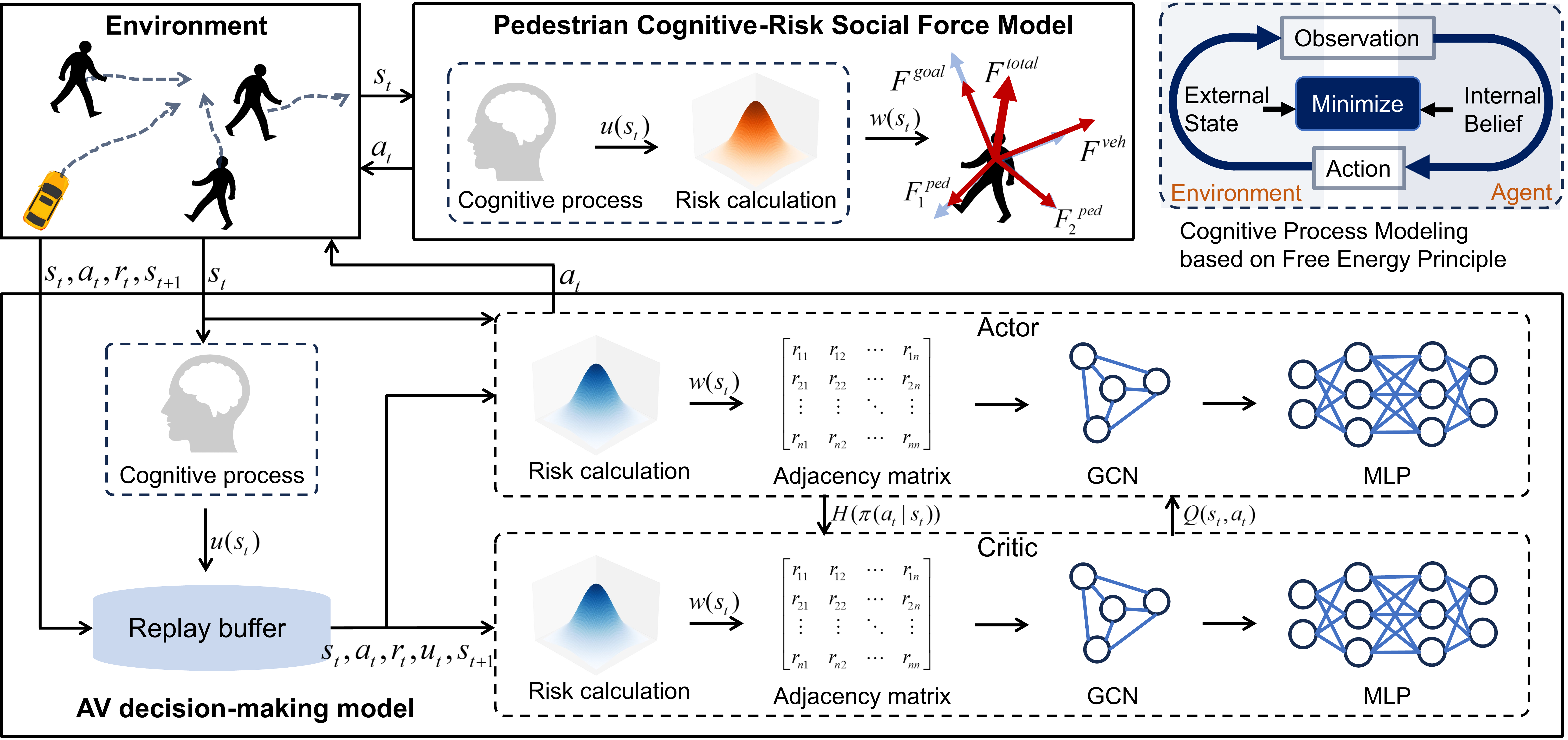}
\caption{The proposed framework of AV-pedestrian interaction in the urban shared space.}
\label{framework}
\end{figure*}
To address these limitations, our work proposes a graph-enhanced deep reinforcement learning framework for modeling interactions between an AV and multiple pedestrians. Fig. \ref{framework} illustrates the overall architecture of our proposed framework in urban shared spaces. In this work, we adopt the Soft Actor-Critic (SAC) algorithm to train the AV’s decision-making policy. Specifically, a Graph Convolutional Network (GCN) module is integrated into both the actor and critic networks to capture structured multi-agent interactions. The GCN operates on a dynamically constructed interaction graph, whose adjacency matrix encodes the interaction intensity between agents. The edge weights are computed using a combined risk metric that integrates physical risk and cognitive uncertainty, with the latter derived from the Free Energy Principle. Furthermore, pedestrians are modeled as intelligent agents with both interactive capabilities and cognitive awareness. Their behavior is generated through the proposed Cognitive-Risk Social Force Model (CR-SFM), which employs the same integrated risk signal used in the GCN to modulate the contribution of the goal-directed and repulsive forces. By incorporating both physical and cognitive risk into the force computation, pedestrian trajectories can be adaptively adjusted in response to surrounding dynamics, yielding more realistic and human-like behaviors in complex environments. \par
In order to better understand the interactions and resolve navigation challenges in crowded environments, the main contributions of this work are as follows: 1). A cognitive uncertainty modeling approach based on the free energy principle is introduced to characterize subjective human prediction biases and the iterative adjustment process when reasoning about the behaviors of other agents. 2). A pedestrian model called the Cognitive-Risk Social Force Model (CR-SFM) is proposed, which dynamically adjusts the weights of goal-directed and repulsive forces based on a fused risk measure. This enables pedestrian models to autonomously adapt their motions in uncertain and interactive scenarios, achieving more realistic behavior than conventional models with fixed coefficients. 3). A risk-encoded interaction graph is constructed to improve AV decision-making within a reinforcement learning framework. The combined risk is used to define the adjacency matrix of a GCN, which is embedded in both the actor and critic networks of the SAC framework. This integration allows the policy to better perceive interactive relationships and potential threats, thus improving decision-making performance. \par
The remainder of this paper is organized as follows: Section \uppercase\expandafter{\romannumeral2} provides a review of the state of the art in the relevant domain. Section \uppercase\expandafter{\romannumeral3} presents the problem statement. Section \uppercase\expandafter{\romannumeral4} details the proposed methodology. The experimental results along with an in-depth analysis are discussed in Section \uppercase\expandafter{\romannumeral5}. Finally, Section \uppercase\expandafter{\romannumeral6} concludes with key findings and potential future work. \par

\section{Related work}
\subsection{Pedestrian Behavior Modeling}
Many pedestrian modeling approaches have been proposed to facilitate the simulation and analysis of human-vehicle interactions in urban shared spaces. Early approaches often rely on simplified pedestrian models with ideal behavioral assumptions that significantly deviated from real-world scenarios. For instance, \cite{r1} modeled pedestrians as moving at a constant velocity, ignoring interactions with the vehicle or other pedestrians. More works have attempted to incorporate interaction dynamics. The study \cite{r2} used recorded pedestrian trajectories, which inherently captured interactions between humans, to train AV policies. However, these trajectories remain static during simulation and do not adapt to the AV’s behavior in real time. Similarly, data-driven pedestrian predictors trained solely on pedestrian datasets \cite{r3} could learn interactions between pedestrians, but failed to account for pedestrian reactions to AV maneuvers due to the absence of vehicle information in their training. Although these models are computationally efficient, they treat pedestrians as passive agents rather than responsive actors, making it difficult to simulate realistic and bidirectional interactions in dynamic environments. \par
Pedestrian reactive models, such as the Optimal Reciprocal Collision Avoidance (ORCA) \cite{r4, r5} or Social Force Model (SFM) \cite{r6, r7, r29}, incorporate agent interactions and collision avoidance capabilities. However, these approaches capture only low-level geometric interactions and overlook higher-level cognitive processes such as risk perception and human behavioral uncertainty, which are critical for robust decision-making in human-centric environments. Therefore, to bridge this gap, we propose the pedestrian Cognitive-Risk Social Force Model (CR-SFM), integrating the joint influence of physical risk and cognitive uncertainty on pedestrian decision-making process. \par

\subsection{AV Decision-making Model}
Due to the limited research on AV navigation in dense pedestrian environments, many insights have been drawn from crowd navigation studies on mobile robots, whose interaction modeling techniques can offer useful inspiration for AV systems. Prior work in this domain falls into two categories: prediction-based planning or learning-based approaches.  \par
Prediction-based planning methods include decoupled and coupled approaches. Decoupled methods first predict the surrounding agents’ trajectories and then plan a collision-free path in the remaining free space \cite{r8}. However, they often lead to conservative behavior or freezing, as the predicted area may become overly occupied with no feasible path \cite{r9}. Coupled methods jointly plan trajectories for all agents by considering mutual interactions \cite{r10}, but they require frequent updates when agents deviate from their planned paths, resulting in high computational cost and limited real-time applications \cite{r11}. Moreover, all prediction-based methods depend heavily on accurate pedestrian trajectory forecasts, which are often unreliable due to the unpredictability of human behavior. \par
Learning-based methods, particularly those based on deep reinforcement learning (DRL), have emerged as promising alternatives due to their ability to capture complex social interactions and support efficient online inference. While most DRL-based crowd navigation studies focus on small mobile robots, only a few have explored AV navigation in dense pedestrian environments. For instance, \cite{r2} trained an AV using a DRL framework with a reward function designed to promote safe and efficient behaviors in crowds. Although this method considers prediction uncertainty, it leverages uncertainty only in the reward design to guide policy learning, without directly incorporating it into the interaction modeling process.\par
To improve interaction modeling, many DRL frameworks for robot navigation have integrated advanced network architectures such as recurrent neural networks (RNN) \cite{r12, r13, r14}, graph neural networks (GNN) \cite{r15,r16,r17}, and attention mechanisms \cite{r18, r19, r20}. For example, \cite{r14} incorporated Long Short-Term Memory into DRL to encode observations from a variable number of surrounding agents into a fixed-length vector. A heterogeneous relational DRL framework was introduced to explicitly model diverse interaction types using a customized GNN \cite{r16}. In \cite{r19}, a spatial graph was combined with an attention module to highlight the relative importance of nearby agents. \par
While these approaches enhance interaction modeling, they often neglect the subjective uncertainty inherent in human cognition, such as prediction errors or biases that may arise when an agent attempts to infer pedestrian intentions or behaviors. To address this, our method adopts a graph-based DRL framework that uniquely encodes a unified representation of physical risk and cognitive uncertainty into an interaction graph. This allows the policy to capture complex and dynamic interactions and make more informed decision-making in pedestrian-rich environments. \par

\section{Problem Formulation}
This work focuses on the AV navigation problem in environments with multiple pedestrians. The objective is to develop a continuous decision-making strategy that ensures safe and efficient operation amid dynamic human interactions. \par
We model this problem as a Markov Decision Process (MDP), defined by the tuple: $\langle \mathcal{S}, \mathcal{A}, \mathcal{P}, \mathcal{R}, \gamma \rangle$. The state space $\mathcal{S}$ captures the dynamic state of the environment at each time step $t$, denoted by \( s_t \in \mathcal{S} \). We assign index $0$ to the AV and indices $1$ to $n$ to the surrounding pedestrians. The AV’s state $s_t^0$ includes its current position $(x, y)$, heading angle $\theta$, velocity $v$, acceleration $\alpha$, goal position $(x_\text{{goal}}, y_{\text{goal}})$, distance to the goal position $d_{\text{goal}}$, and angular deviation from the goal direction $\Delta\theta_{\text{goal}}$. Each pedestrian’s state $s_t^i$, for $i = 1 … n$, consists of their position, heading angle, and walking velocity. The whole environment state at time $t$ is defined as $s_t$ : $(s_t^0, s_t^1, \dots, s_t^n)$. The action space $\mathcal{A}$ represents the set of possible actions available to the AV. Each action $a_t \in \mathcal{A}$, is defined as a tuple $a_t = [\alpha, \Delta\theta]$, where $\alpha$ denotes the acceleration and $\Delta\theta$ denotes the change in heading direction. The state transition probability $\mathcal{P}(s_{t+1}|s_t, a_t)$ defines how the environment evolves from the current state $s_t$ to next state $s_{t+1}$ given the action taken by the AV. Following the transition, the AV receives a reward $r_t \in \mathcal{R}$, indicating the immediate utility of its chosen action. \par
The AV model aims to learn a policy $\pi(a_t|s_t)$ that produces a sequence of optimal actions to maximize the expected cumulative discounted reward over time, where the process terminates upon reaching the destination. This decision-making problem can thus be defined as follows:
\begin{equation}
\begin{aligned}
    & \underset{\pi}{\text{max}}
    & & \mathbb{E} \left [\sum_{t=0}^{\infty} \gamma^t r_t(s_t, a_t) \mid a_t\sim\pi  \right] \\
    & \text{subject to}
    & & s_{t+1} \sim \mathcal{P}(s_{t+1} \mid s_{t}, a_{t}), \\
    & & & a_t \in \mathcal{A}, s_t \in \mathcal{S}, \\
    & & & r_{t} \in \mathcal{R}, \forall t \in \mathbb{N}
\end{aligned}
\end{equation}
where $\gamma \in [0, 1]$ is the discount factor that determines the relative importance of future rewards.

\section{Methodology}
\subsection{Cognitive Uncertainty Modeling}
One of the core computational tasks of the human brain is to infer latent world states and predict future behaviors of others or the environment based on sensory inputs and prior knowledge, especially in uncertain and dynamic conditions \cite{r31}. In this process, individuals often experience a lack of full confidence in their predictions. This subjective uncertainty, referred to as cognitive uncertainty \cite{r33}, reflects how confident an individual is about expected outcomes \cite{r34}. When predictions deviate from actual observations, the resulting errors drive belief adjustments and expectation updates, leading to more rational decisions. For example, a pedestrian intending to cross the road may expect an approaching vehicle to slow down and yield. If the pedestrian lacks confidence in this expectation, it reflects a high level of cognitive uncertainty. If the vehicle accelerates instead, the mismatch prompts the pedestrian to re-evaluate and adjust their behavior. Capturing such cognitive processes is vital for modeling human decision-making under uncertainty. Integrating this cognitive mechanism into intelligent agent models allows for more human-like and interpretable behavior, thereby enhancing the realism of human–machine interaction. \par
The Free Energy Principle (FEP) provides a unifying theoretical framework for describing this cognitive process \cite{r32, r35}. It suggests that agents continuously generate predictions and revise internal beliefs based on sensory feedback in order to minimize “surprise” and maintain cognitive stability and behavioral adaptability. In Bayesian Inference, “surprise” is typically formalized as the negative log marginal likelihood, $- \log P(o)$. Since the marginal likelihood $P(o) = \int p(x, o)\, \text{d}x$ is generally intractable, direct minimization is impractical. To overcome this, the FEP introduces variational free energy as an optimizable upper bound \cite{r36}:
\begin{equation}
F(Q) = D_{\mathrm{KL}}(Q(x) \| P(x|o)) - \log P(o)
\end{equation}
where $Q(x)$ denotes the agent’s approximate posterior belief, and $P(x|o)$ is the true posterior given the observation $o$. The Kullback-Leibler (KL) divergence quantifies the discrepancy between the two distributions. Since the KL divergence is always non-negative, minimizing free energy pushes the internal belief distribution toward the true posterior. This process provides a mathematical foundation for cognitive modeling. \par
In practical scenarios, cognitive uncertainty, as a subjective mental state, is inherently difficult to observe or measure directly. Individuals often lack confidence in their judgments when exposed to limited information or complex environments. In order to quantify this uncertainty, we propose the following modeling assumptions: when the discrepancy between predicted and observed outcomes is large, the individual's confidence in their prediction is low, indicating higher cognitive uncertainty. Accordingly, we adopt the KL divergence between the predicted and the observed distributions as a proxy measure of epistemic uncertainty, capturing the agent’s confidence level in forecasting the future behavior of others. \par
In this work, we use velocity as the target variable and implement a closed-loop cognitive model based on the predict–observe–update process, grounded in the Free Energy Principle and Bayesian Inference. Each agent’s prediction of others’ velocities is assumed to follow a Gaussian distribution. At each time step $t$, the ego agent, whether an AV or a pedestrian, generates a velocity prediction $\hat{v}_t$ for surrounding agents, and combines it with the posterior from the previous step $t-1$ to obtain the current prior:
\begin{equation}
p_t(v) = \mathcal{N}(\mu_p, \sigma_p^2)
\end{equation}
Upon observing the actual velocity $v_t$, an observation distribution is constructed with the observed value as its mean $\mu_o$ and a fixed variance $\sigma_o^2$:
\begin{equation}
o_t(v) = \mathcal{N}(\mu_o, \sigma_o^2)
\end{equation}
The cognitive uncertainty is approximated by computing the KL divergence between the predicted and observed distributions, expressed as:
\begin{equation}
u_t = D_{KL}(p_t(v)||o_t(v))
\end{equation}
A larger KL value indicates a greater deviation, implying lower confidence in the prediction and higher epistemic uncertainty. The ego agent then performs Bayesian updating to obtain the posterior \cite{r37}:
\begin{equation}
\frac{1}{\sigma_{\text{post}}^2} = \frac{1}{\sigma_p^2} + \frac{1}{\sigma_o^2}
\end{equation}
\begin{equation}
\quad\mu_{\text{post}} = \sigma_{\text{post}}^2 \left( \frac{\mu_p}{\sigma_p^2} + \frac{\mu_o}{\sigma_o^2} \right)
\end{equation}
This posterior represents the agent’s revised belief about others’ behavior at the current time step and serves as the prior for the next prediction, thus forming a recursive cognitive inference process. The complete procedure is presented in Algorithm \ref{alg1}.
\begin{algorithm}[htbp]
\caption{Human cognitive process modeling with cognitive uncertainty quantification}
\label{alg1}
\KwIn{Surrounding agents set $N$, time horizon $T$, internal prediction model}
\KwOut{Cognitive uncertainty}
\For{each time step $t \in T$}{
    \For{each surrounding agent $i \in N$}{
        Use internal prediction model to predict velocity $\hat{v}_t^i$; \\
        Obtain prior distribution $p_t^i(v)$ based on predicted velocity $\hat{v}_t^i$ and posterior $p_{\text{post},t-1}^{i}(v)$; \\
        Observe true velocity $v_t^i$ and construct observation distribution $o_t^i(v)$; \\
        Compute cognitive uncertainty $u_t^{i}$; \\
        Compute posterior distribution $p_{\text{post},t}^{i}(v)$ using Bayesian update; \\
        Store the updated posterior $p_{\text{post},t}^{i}(v)$; \\
    }
}
\end{algorithm}

\subsection{Pedestrian Cognitive-Risk Social Force Model}
The social force model \cite{r40} describes pedestrian movement as the result of a combination of virtual social forces, including a goal force that drives individuals toward their destinations, and repulsive forces from other agents or obstacles to avoid collisions. These forces are represented as vectors and superimposed to determine the pedestrian’s overall motion, enabling modeling and simulation of pedestrian behavior. In the AV–pedestrians interaction considered in this study, each pedestrian is influenced by three primary forces: a goal force guiding movement toward the destination, a vehicle-induced repulsive force from the AV, and repulsive forces from surrounding pedestrians. \par
To model efficient goal-oriented movement, the goal force for pedestrian $i$ is defined as:
\begin{equation}
\mathbf{F}_i^{\text{goal}} = \frac{v_i^0 \vec{e}_i - \vec{v}_i}{\tau_i}
\end{equation}
where $v_i^0$ is the desired speed, $\vec{e}_i$ is the unit vector pointing toward the destination, $\vec{v}_i$ is the current velocity of pedestrian $i$, and $\tau_i$ is the relaxation time representing how quickly the pedestrian adapts to the desired velocity.
The repulsive interactions are modeled as exponentially decaying functions of distance. The repulsive force from the AV is given by:
\begin{equation}
\mathbf{F}_i^{\text{veh}} = A_{\text{veh}} \cdot \exp\left( \frac{r_{i,\text{veh}} - d_{i,\text{veh}}}{B_{\text{veh}}} \right) \cdot \vec{n}_{i,\text{veh}}
\end{equation}
and the repulsive force from another pedestrian $j$ is computed as:
\begin{equation}
\mathbf{F}_{ij}^{\text{ped}} = A_{ij} \cdot \exp\left( \frac{r_{ij} - d_{ij}}{B_{ij}} \right) \cdot \vec{n}_{ij}
\end{equation}
In both equations, $A$ and $B$ are constants; $d$ is the Euclidean distance between agents; $r$ is the sum of physical radii; and $\vec{n}$ is the unit vector pointing from the interacting agent to pedestrian $i$.
These forces are combined to yield the resultant force acting on each pedestrian. For pedestrian $i$, the total force can be formulated as \cite{r41}:
\begin{equation}
\label{nsfm}
\mathbf{F}_i = \mathbf{F}_i^{\text{goal}} + \mathbf{F}_i^{\text{veh}} + \sum_{\substack{j \in \mathcal{G} \\ j \ne i}} \mathbf{F}_{ij}^{\text{ped}}
\end{equation}
where $\mathcal{G}$ denotes the set of all pedestrians in the current scene. \par
While the social force model has been widely applied in pedestrian behavior modeling, it still shows certain limitations in complex human-vehicle interaction scenarios. On the one hand, the magnitude of repulsive forces is usually fixed and based solely on spatial distance, without accounting for the velocity and motion direction of interacting agents. For instance, an AV may be physically close to a pedestrian but moving away; in such cases, the risk is negligible. However, the traditional model would still exert a strong repulsive force purely due to proximity, leading to unrealistic or overly cautious avoidance behaviors. On the other hand, the model lacks a representation of  human cognitive processes. In real-world interactions, pedestrians continuously perceive, anticipate, and update their beliefs about others’ behaviors. This “perception–prediction–observation–update” loop is fundamental to human decision-making in dynamic, uncertain environments. But it is absent in the traditional social force model.\par
\begin{figure}[htp]
\centering
\begin{subfigure}[b]{0.49\linewidth}
        \centering
        \includegraphics[width=\linewidth]{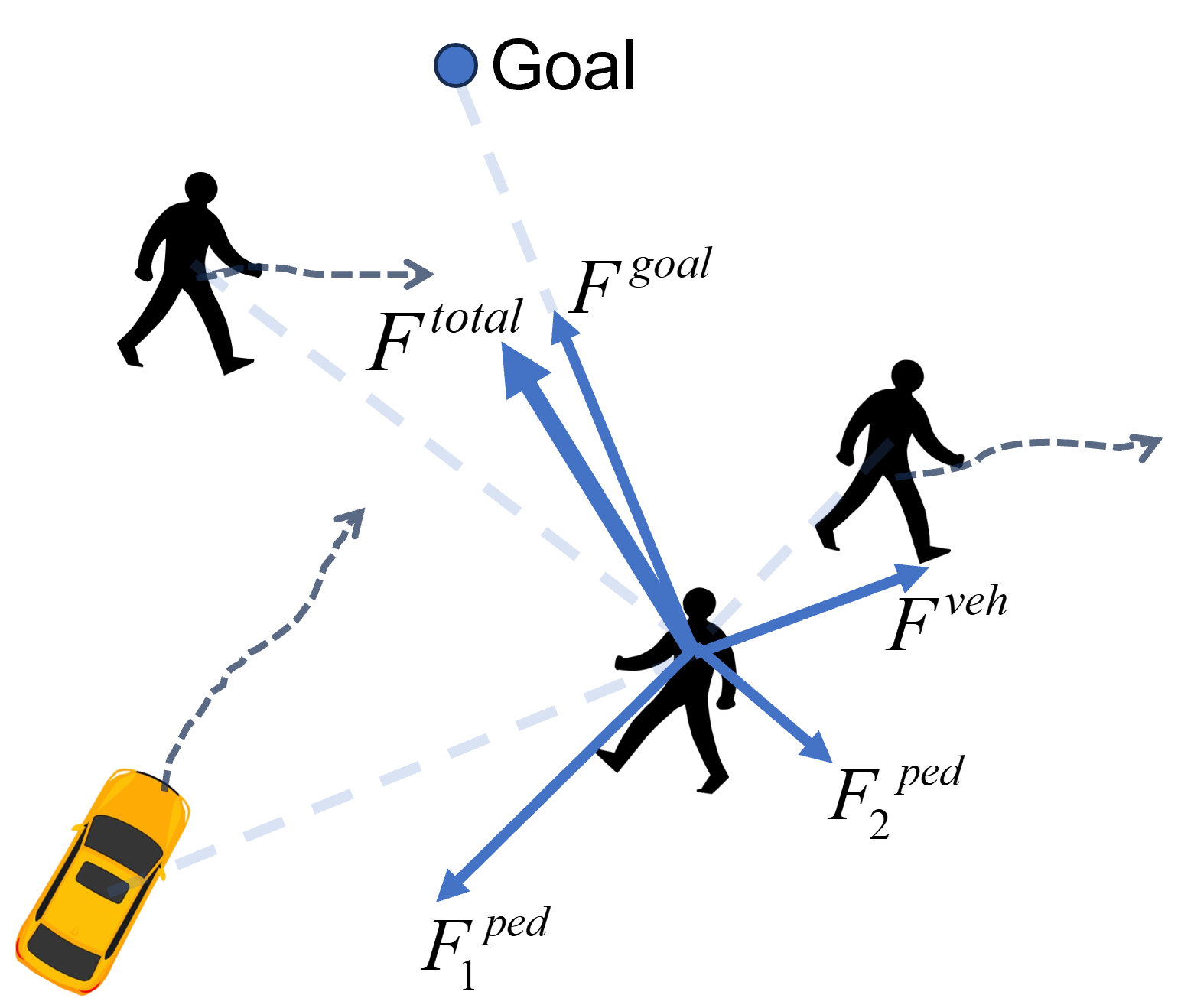}
        \caption{}
    \end{subfigure}
    \hfill
    \begin{subfigure}[b]{0.49\linewidth}
        \centering
        \includegraphics[width=\linewidth]{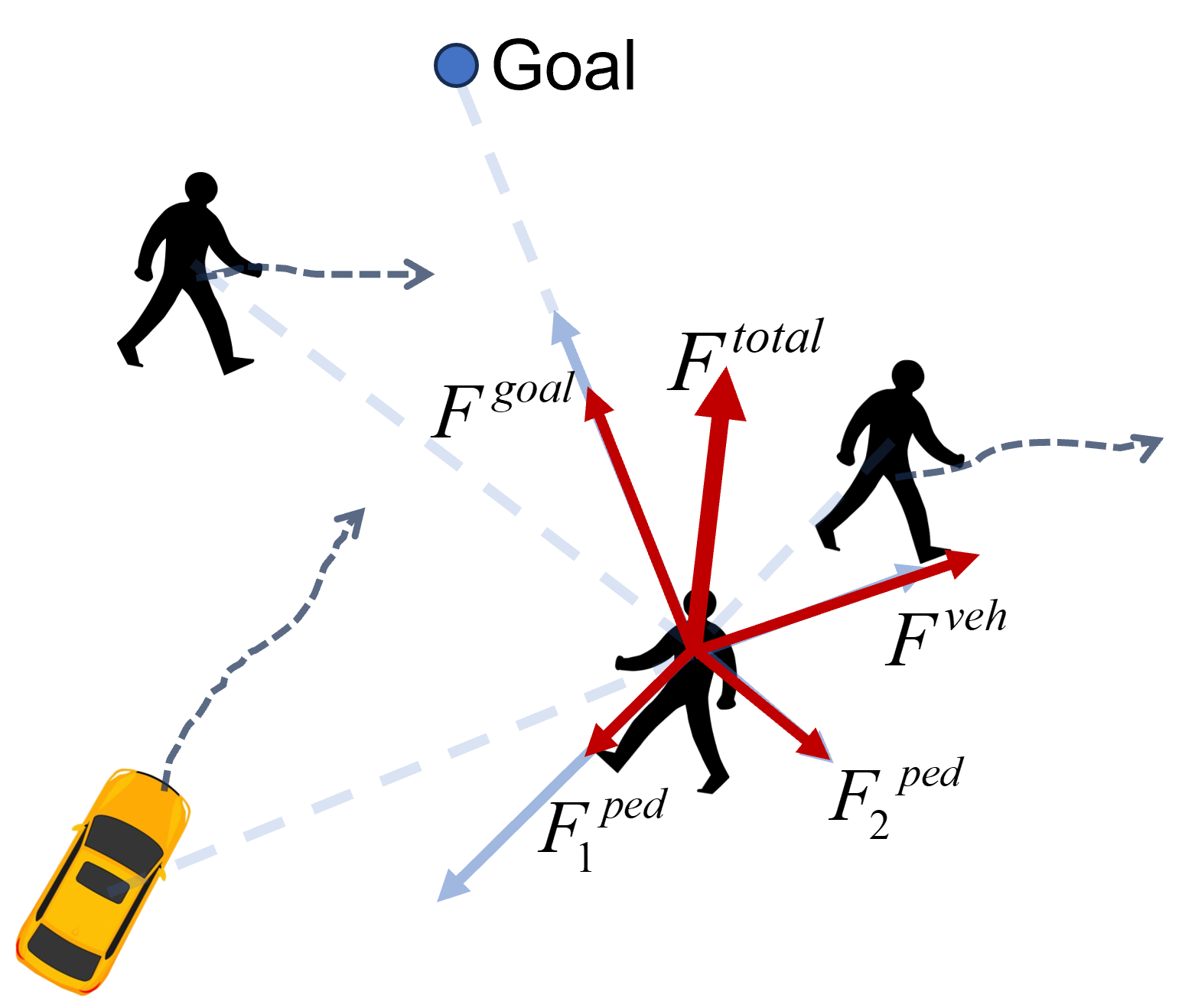}
        \caption{}
    \end{subfigure}
\caption{Comparison of the SFM and CR-SFM in pedestrian-vehicle interactions. These figures show the force composition acting on the ego pedestrian under the SFM (a) and CR-SFM (b). In the SFM, all repulsive forces are distance-based with fixed weights. Pedestrian 1 (the rightmost pedestrian in the figure), being closest, produces the strongest repulsion, shifting the total force toward the AV and increasing collision risk. In CR-SFM, force weights are adjusted based on physical features (e.g., distance, velocity, direction) and cognitive uncertainty. Although pedestrian 1 is nearest, his motion away leads to low perceived risk and a weaker force. The approaching AV contributes more strongly, shifting the total force toward a safer direction.}
\label{socialforce}
\end{figure}

To overcome these limitations, we propose the pedestrian Cognitive-Risk Social Force Model (CR-SFM). The core idea is to introduce risk weight coefficients that dynamically modulate the intensity of each force, allowing pedestrian behaviors to adapt responsively to evolving interaction contexts. Fig. \ref{socialforce} provides a schematic comparison between the SFM and CR-SFM, showing the difference in resultant force under the same interaction scenario. The revised resultant force acting on pedestrian $i$ is formulated as:
\begin{equation}
\mathbf{F}_i = w_i^{\text{goal}} \cdot \mathbf{F}_i^{\text{goal}} + w_i^{\text{veh}} \cdot \mathbf{F}_i^{\text{veh}} + \sum_{\substack{j \in \mathcal{G} \\ j \ne i}} w_{ij}^{\text{ped}} \cdot \mathbf{F}_{ij}^{\text{ped}} 
\end{equation}
where $w_i^{\text{goal}}$, $w_i^{\text{veh}}$ and $w_{ij}^{\text{ped}}$ represent the risk-based weights for the goal force, vehicle repulsion, and pedestrian repulsion, respectively. These weights are computed by combining physical risk and cognitive uncertainty. \par
In dynamic interaction scenarios, risk assessment methods based solely on geometric distance often fail to accurately capture the actual level of threat, as they overlook the critical influence of relative motion on risk perception. For example, at the same distance, an agent approaching at high speed poses a greater danger than one moving slowly or moving away. To better reflect this perceptual difference, we introduce a risk modeling approach based on virtual distance. By incorporating dynamic motion features such as velocity, acceleration, and direction \cite{r38} to modulate the actual distance, the resulting virtual distance is then used to replace the original geometric distance for physical risk computation. The physical risk $\psi$ is defined as:
\begin{equation}
d_\text{virtual} = d_{\text{actual}} \cdot \left(1 + \tanh\left( \gamma_1 \cdot k \cdot \left(v \cdot |\cos\varphi| + |a| \right) \right) \right)
\end{equation}
\begin{equation}
k = 
\begin{cases}
1, & \text{if the agent is moving away} \\
-1, & \text{if the agent is approaching}
\end{cases}
\end{equation}
\begin{equation}
\psi = \frac{1}{1 + \gamma_2 \cdot d_\text{virtual}}
\end{equation}
where $d_{\text{actual}}$ is the actual physical distance between the pedestrian and the interacting agent, $v$ and $a$ are the velocity and acceleration of the interacting agent, $\varphi$ is the angle between the agent’s velocity vector and the vector pointing from the agent to pedestrian $i$, and $\gamma_1$ and $\gamma_2$ are hyperparameters. These formulations imply that risk increases when an approaching agent moves towards the pedestrian with higher speed or greater acceleration.\par
Cognitive uncertainty captures the pedestrian’s subjective confidence in predicting others’ future behaviors. Rather than being modeled as an independent risk component, it is treated as a modulatory factor that amplifies physical risk under uncertain situations. For example, when physical risk is low, but the pedestrian is unsure of the AV’s future behavior, a higher cognitive uncertainty increases perceived risk, enhancing repulsive forces and encouraging earlier avoidance. 
The risk weight coefficients are calculated as follows: 
\begin{equation}
w_i^{\text{veh}} = \psi_i^{\text{veh}} \cdot \left(1 + \lambda_1 \cdot u_i^{\text{veh}}\right)
\end{equation}
\begin{equation}
w_{ij}^{\text{ped}} = \psi_{ij}^{\text{ped}} \cdot \left(1 + \lambda_2 \cdot u_{ij}^{\text{ped}}\right)
\end{equation}
\begin{equation}
w_i^{\text{goal}} = \exp\left(-\lambda_3 \cdot \max\left(w_i^{\text{veh}}, \max_{j \in \mathcal{G}, j \neq i} w_{ij}^{\text{ped}}\right)\right)
\end{equation}
where $u_i^{veh}$ and $u_{ij}^{ped}$ denote cognitive uncertainties. $\lambda_1$, $\lambda_2$, $\lambda_3$ are scaling hyperparameters. Notably, the goal force weight is modeled using an exponential decay function to reduce goal-oriented behavior when perceived interaction risk is high, allowing the pedestrian to prioritize avoidance over forward motion. \par
In summary, by introducing a dynamic weighting mechanism that integrates both physical and cognitive factors, the CR-SFM can make the pedestrian to modify their approach and avoidance strategies based on perceived risk and uncertainty, more closely aligning with real human decision-making processes.  \par

\subsection{Graph-Enhanced DRL for AV Decision-Making }
In this study, we adopt the Soft Actor-Critic (SAC) as a basic reinforcement learning framework to guide the AV decision-making strategies in complex, pedestrian-rich environments. The SAC is grounded in the Maximum Entropy Reinforcement Learning theory. While optimizing the expected cumulative return, it explicitly introduces a policy entropy term to encourage greater stochasticity and exploration under uncertainty. The objective is to maximize the following expression \cite{r45}:
\begin{equation}
J(\pi) = \sum_{t} \mathbb{E}_{(s_t, a_t) \sim \pi} \left[ r(s_t, a_t) + \alpha \mathcal{H}(\pi(\cdot | s_t)) \right]
\end{equation}
where $\mathcal{H}(\pi(\cdot | s_t))$ is the entropy of the policy at state $s_t$, and $\alpha$ is a temperature coefficient balancing reward and entropy. \par
The SAC follows an actor-critic architecture consisting of a policy network (actor), two Q-value networks (critics), and their corresponding target networks. The actor outputs a Gaussian distribution over actions given a state $s_t$, from which an action $a_t$ is sampled. The Q-networks estimate the value of the state-action pair $(s_t, a_t)$. Using samples from a replay buffer, the networks are updated iteratively to approximate optimal behavior. To mitigate Q-value overestimation, SAC employs a double Q-network structure, where the minimum of the two Q-values is used to compute the target value. The critic is trained to minimize the following loss:
\begin{equation}
\mathcal{L}_{\text{critic}} = \mathbb{E}_{(s_t, a_t, r_t, s_{t+1}) \sim \mathcal{D}} \left[\frac{1}{2} \left( Q(s_t, a_t) - y_t \right)^2 \right]
\end{equation}
with the target value $y_t$ defined as:
\begin{equation}
y_t = r_t + \gamma \min_{i=1,2} Q_i'(s_{t+1}, a_{t+1}) - \alpha \log \pi(a_{t+1} \mid s_{t+1})
\end{equation}
The target Q-networks are updated softly via:
\begin{equation}
\theta' \leftarrow \tau \theta + (1 - \tau) \theta'
\end{equation}
where $\tau \in (0, 1)$ is a small soft update coefficient, $\theta$ denotes the parameters of the current critic network, and $\theta'$ denotes the parameters of the target critic network.
The actor is trained to maximize the expected Q-value while maintaining high entropy:
\begin{equation}
\mathcal{L}_{\text{actor}} = \mathbb{E}_{s_t \sim \mathcal{D}} \left[ \alpha \log \pi(a_t|s_t) - Q(s_t, a_t) \right]
\end{equation}
Through this maximum entropy optimization framework, the SAC enables stable and continuous policy learning in complex environments, allowing the AV to develop strategies that are both efficient and exploratory in dynamic multi-agent interactions. \par
However, the standard SAC architecture does not explicitly model dynamic interactions among multiple agents, which is critical in autonomous driving scenarios involving dense pedestrian. To address this, we incorporate a Graph Convolutional Network (GCN) into the SAC framework to extract interaction-aware state representations from the multi-agent graph. \par
At each time step, the AV and surrounding pedestrians are modeled as graph nodes, with the adjacency matrix encoding their interactions. Given a graph $\boldsymbol{G}_t=(\boldsymbol{V}_t,\boldsymbol{A}_t)$, where $\boldsymbol{V}_t$ is a set of all agents at time $t$, $\boldsymbol{A}_t$ is the adjacency matrix, $\boldsymbol{H}^{(l)} \in \mathbb{R}^{N \times d}
$ represents the node feature at layer $l$, and the GCN propagation rule is expressed as \cite{r39}:
\begin{equation}
\boldsymbol{H}^{(l+1)} = \sigma \left( \boldsymbol{\widetilde{D}}^{-1/2} \boldsymbol{\widetilde{A}} \boldsymbol{\widetilde{D}}^{-1/2} \boldsymbol{H}^{(l)} \boldsymbol{W}^{(l)} \right)
\end{equation}
where $\boldsymbol{\tilde{A}} = \boldsymbol{A} + \boldsymbol{I}$ is the adjacency matrix adding self-connection, $\boldsymbol{\tilde{D}}$ is the degree matrix, $\boldsymbol{W}^{(l)}$ is the learnable weight matrix, and $\sigma$ is the nonlinear activation function. This operation aggregates each agent’s state with information from its neighbors, capturing local structural relationships and interaction patterns. The resulting high-order representations serve as enhanced state features for downstream policy learning and value estimation. \par
However, the adjacency matrix in conventional GCNs is often constructed using static graph topology, making it insufficient for modeling dynamic and asymmetric interaction patterns over time. In real-world multi-agent interactions, humans do not pay equal attention to all agents, but instead focus on potentially high-risk individuals that are closer, faster, or have unpredictable behaviors. \par
Inspired by this, we propose a risk encoded interaction graph, where physical risk and cognitive uncertainty are combined into a unified risk score. This value directly determines the attention weights in the adjacency matrix, allowing the model to focus more effectively on high-risk interactions during information aggregation. Specifically, we construct a dynamic adjacency matrix $\boldsymbol{A}_t$ at each time step, where each element $A_t(i, j)$ represents the influence of agent $j$ on agent $i$:
\begin{equation}
A_t(i, j) = \psi_t(i, j) \cdot \left( 1 + \lambda \cdot u_t(i, j) \right)
\end{equation}
where $\psi_t(i, j)$ denotes the physical risk based on agent $j$'s relative position, speed, and direction with respect to agent $i$, while $u_t(i, j)$ quantifies agent $i$'s uncertainty about agent $j$'s behavior, computed using the free-energy-based method described in Algorithm \ref{alg1}. $\lambda$ is a hyperparameter controlling the influence of uncertainty. \par
This dynamic adjacency matrix guides GCN propagation to focus on high-risk interactions, enabling the model to emphasize critical agents and filter out irrelevant ones. Compared to traditional graph modeling, our approach jointly encodes physical structure and cognitive cues, providing the policy network with more decision-relevant state representations. This enhances the policy's ability to identify key interactions and make robust decisions in complex, multi-agent environments. The overall training process is summarized in Algorithm~\ref{alg2}.

\begin{algorithm}[htbp]
\caption{SAC for AV Decision-Making with Cognitive Process}
\label{alg2}
Initialize critic networks $Q_1$, $Q_2$, actor network $\pi$, target networks $Q_1'$, $Q_2'$, and replay buffer $\mathcal{D}$\;
\For{episode $i \in M$}{
    Observe initial state $s_0$\;
    \For{each time step $t \in T$}{
        Compute cognitive uncertainty $u_t$ using Algorithm~\ref{alg1}\;
        Execute an action $a_t$, receive a reward $r_t$, and observe the next state $s_{t+1}$\;
        Store the transition $(s_t, a_t, r_t, s_{t+1}, u_t)$ in replay buffer $\mathcal{D}$\;
        Update $s_t \gets s_{t+1}$\;
        Sample a minibatch of transitions from $\mathcal{D}$\;
        Compute target Q-value $y_t$\;
        Update critics $Q_1$ and $Q_2$ by minimizing loss $\mathcal{L}_{\text{critic}}$\;
        Update actor network $\pi$\;
        Soft update target critic networks $Q_1'$ and $Q_2'$\;
    }
}
\end{algorithm}

\section{Experiment}
\subsection{Experimental Setup}
To evaluate the effectiveness of the proposed pedestrian CR-SFM and the AV decision-making model, we conducted a series of simulation experiments in a shared-space interaction scenario. \par
We adopted the real-world Hamburg Bergedorf Station (HBS) dataset \cite{r42}, which captures a typical shared-space street environment in Hamburg, Germany. This dataset contains rich interactions between low-speed vehicles and multiple crossing pedestrians in unconstrained urban settings. From the HBS dataset, we extracted 78 representative interaction scenarios, each involving a single AV and three pedestrians. These scenarios were divided into a training set (77\%) and test set (23\%) to support model learning and performance evaluation, respectively. Each simulation was initialized using the recorded initial positions and velocities of all agents. During the interaction process, the AV generated its actions based on the learned policy, while pedestrian behaviors were governed by our proposed CR-SFM. A simulated episode terminated when the AV reached its goal, collided with a pedestrian, or the maximum time horizon was exceeded. \par
All experiments were conducted on a high-performance laptop equipped with an Intel i9 processor and an NVIDIA RTX 4090 GPU. Model training and simulation were implemented using PyTorch in a custom-built environment. Due to computational limitations on local hardware, we restricted our simulation setup to include three pedestrians interacting with an AV. This configuration balances training feasibility and interaction complexity, while still capturing key social dynamics observed in shared-space environments. In the simulation, the AV's maximum speed was set to 6 m/s, with acceleration constrained to the range of [-2, 2] m/s$^2$. Pedestrians were limited to a maximum walking speed of 2 m/s. The simulation operated with a discrete time step of 0.5 seconds. 

\subsection{Comparative Models and Evaluation Metrics}
This section describes the comparative models as well as evaluation indicators used for both the pedestrian model and the AV decision-making model. \par
For the pedestrian model, in order to assess the performance of our proposed pedestrian CR-SFM, we conducted comparative experiments with two baseline models and one ablation variant. The compared models are detailed as follows: \par
\textit{Constant Velocity Model (CV)} \cite{r1}: In this model, the pedestrian maintains a constant initial velocity toward the predefined goal throughout the entire episode. The motion is purely goal-driven and remains unaffected by the presence or behavior of surrounding agents. \par
\textit{Social Force Model (SFM)} \cite{r41}: This classical model serves as the primary baseline. It describes the pedestrian’s motion as the result of three interacting forces, as defined in Equation \ref{nsfm}. In the standard formulation, the magnitudes of these forces are uniformly weighted using fixed coefficients (set to 1), without any dynamic modulation in response to changes in interaction context. \par
\textit{Risk-aware Social Force Model (RA-SFM)}: This variant serves as an ablation model to evaluate the specific contribution of cognitive uncertainty. It extends the standard SFM by introducing physical risk as a modulation factor for social force magnitudes. Specifically, the strengths of the goal force, vehicle repulsion force, and pedestrian repulsion force are dynamically adjusted based on the estimated physical risk from surrounding agents. \par
\textit{Cognitive-Risk Social Force Model (CR-SFM)}: This is our proposed model. Building upon the RA-SFM, CR-SFM incorporates both physical risk and cognitive uncertainty into the modulation of force magnitudes. \par
To ensure a fair comparison, all models are calibrated using the same set of real-world pedestrian trajectories extracted from the training dataset. Bayesian Optimization is employed to fit the model parameters, aiming to best replicate the real pedestrian behaviors. All models are operated under the same experimental conditions, using the following three metrics for trajectory evaluation: \par
\textit{Average Displacement Error (ADE)}: ADE measures the average deviation between the simulated trajectory and the ground truth trajectory across all time steps. For each episode, we compute the mean Euclidean distance between simulated and actual positions across every time step. The final ADE is obtained by averaging these values over all episodes \cite{r43}:
\begin{equation}
\text{ADE} = \frac{1}{M} \sum_{m=1}^{M} \left( \frac{1}{T} \sum_{t=1}^{T} \left\| \hat{x}_t - x_t \right\| \right)
\end{equation}
where $M$ is the total number of test episodes, $T$ is the number of time steps in each episode, $\hat{x}_t$ denotes the simulated position of at time step $t$, and $x_t$ is the corresponding ground-truth position. \par
\textit{Final Displacement Error (FDE)}: FDE evaluates the final-point accuracy by measuring the Euclidean distance between the simulated final position and the true final position \cite{r43}: \par
\begin{equation}
\text{FDE} = \frac{1}{M} \sum_{m=1}^{M} \left\| \hat{x}_{T} - x_{T} \right\|
\end{equation}

\textit{Collision Rate (CR)}: CR is defined as the proportion of simulation episodes in which the simulated pedestrian model collides with other agents. It is calculated as the ratio of collision cases to the total number of episodes. A lower CR value indicates better collision avoidance performance. \par
To evaluated our AV decision-making model, we compare it with one strong baseline and two ablation variants: \par
\textit{Uncertainty-Aware Polar Collision Grid (UAW-PCG)} \cite{r2}: This state-of-the-art DRL framework integrates pedestrian trajectory prediction uncertainty into AV navigation in crowded environments. It employs a Polar Collision Grid predictor to estimate future pedestrian positions along with their associated uncertainty. These predictions are then incorporated into both the input state and the design of the reward function. The AV decision-making policy is optimized using Proximal Policy Optimization (PPO) algorithm. \par
\textit{Standard SAC (S-SAC)}: This baseline uses a standard SAC framework, where both the actor and critic networks consist only of multilayer perceptron without GCN modules. It does not explicitly capture interactions between the AV and surrounding agents, and serves to evaluate the benefit of introducing graph-based structures. \par
\textit{Graph-enhanced SAC without cognitive modeling (G-SAC-NoCog)}: This ablation model adopts the GCN module into the SAC framework to enhance the perception of multi-agent interactions. However, it uses a static adjacency matrix with uniform weights between all agent pairs, ignoring both physical risk and cognitive uncertainty. This model is designed to isolate the contribution of our proposed risk-encoded interaction representation. \par
\textit{Graph-enhanced SAC with cognitive modeling (G-SAC-Cog)}: This is our proposed model. It integrates a GCN module to capture structured multi-agent interactions, with the adjacency matrix dynamically constructed based on a fused risk signal that combines physical risk and cognitive uncertainty. \par
Following the evaluation protocol used in \cite{r2, r44}, we assess each model from the perspectives of safety, efficiency and smoothness. Specifically, safety is quantified using three indicators: the success rate (the proportion of episodes in which the AV reaches its destination without any collisions), the collision rate (the frequency of collisions with surrounding agents), and the timeout rate, which accounts for episodes where the AV fails to reach the goal within the maximum allowed time despite avoiding collisions. Efficiency is evaluated based on the average vehicle speed, which reflects the efficiency of the AV’s navigation performance. Lastly, smoothness is measured by the average jerk and average maximum absolute acceleration/deceleration, capturing the stability and comfort of the AV’s motion during navigation.

\section{Results and discussion}
This section presents the qualitative and quantitative validation results of both the proposed pedestrian CR-SFM and AV decision-making model.
\subsection{Pedestrian Cognitive-Risk Social Force Model}
\subsubsection{Qualitative Analysis}
\begin{figure}[htp]
    \centering
    \begin{subfigure}[b]{0.45\textwidth}
        \centering
        \includegraphics[width=\textwidth]{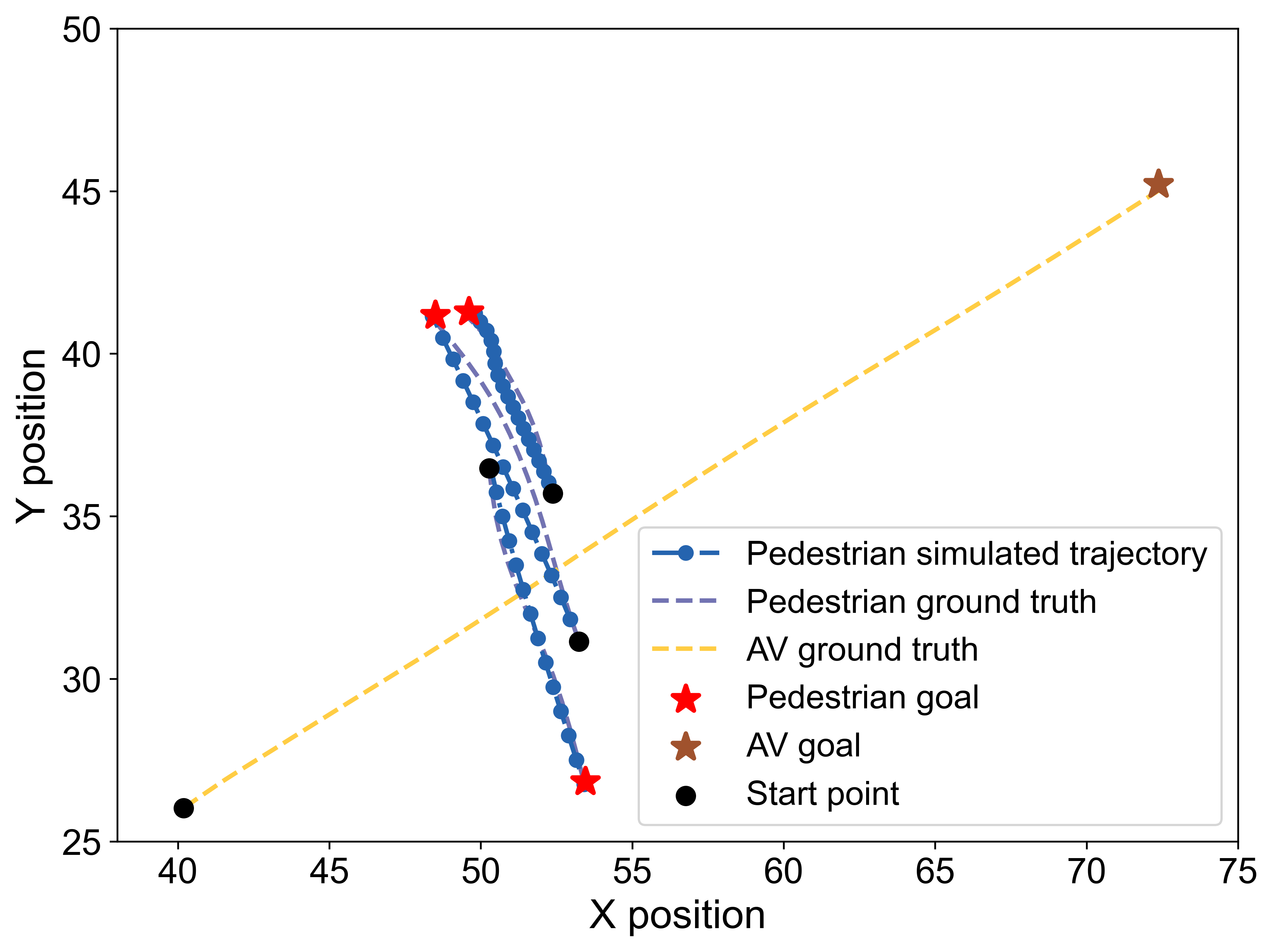}
        \caption{}
        \label{ped1}
    \end{subfigure}
    \vspace{0.5em}
    \begin{subfigure}[b]{0.45\textwidth}
        \centering
        \includegraphics[width=\textwidth]{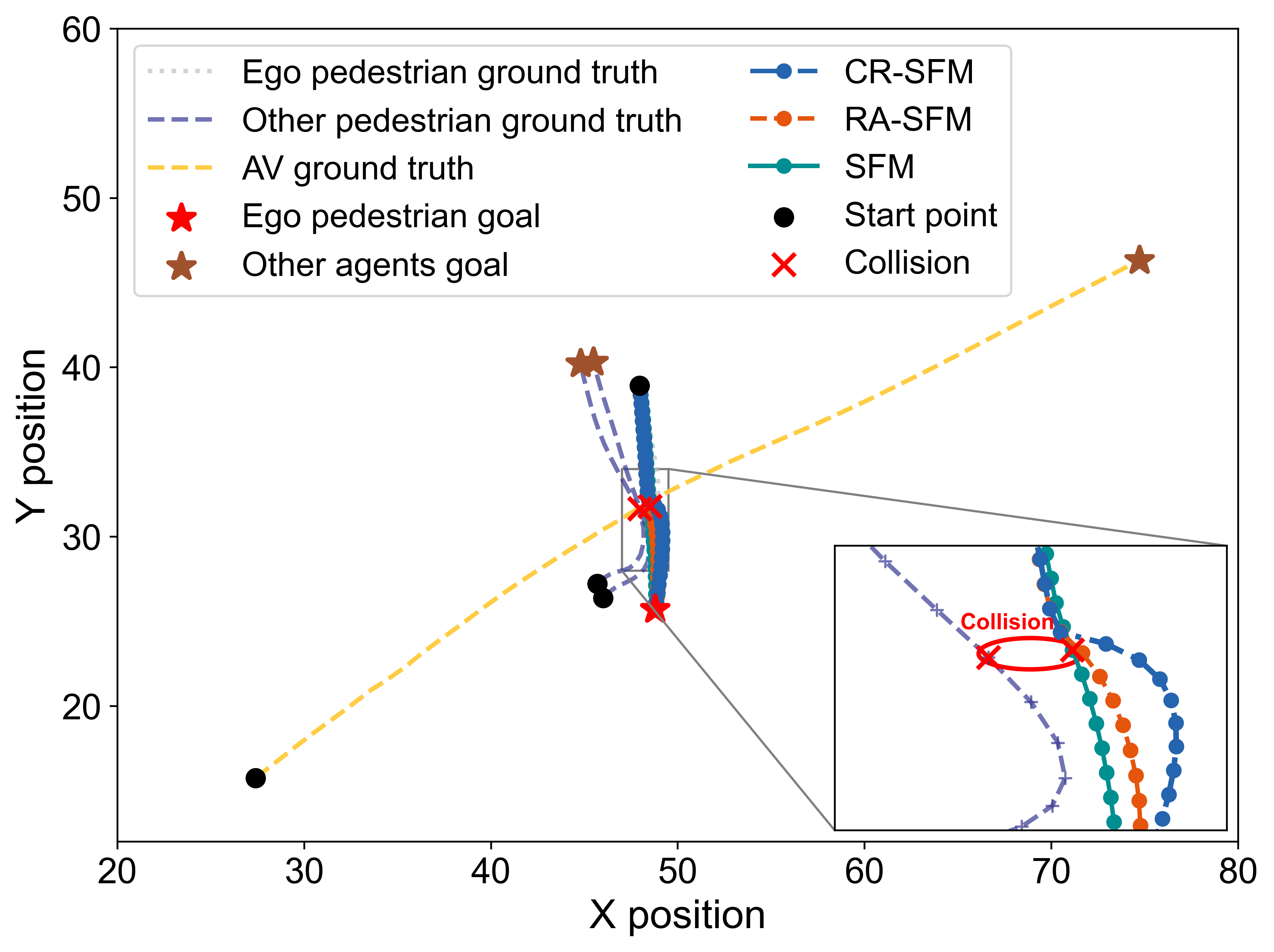}
        \caption{}
        \label{ped2}
    \end{subfigure}
    \vspace{0.5em}
    \begin{subfigure}[b]{0.45\textwidth}
        \centering
        \includegraphics[width=\textwidth]{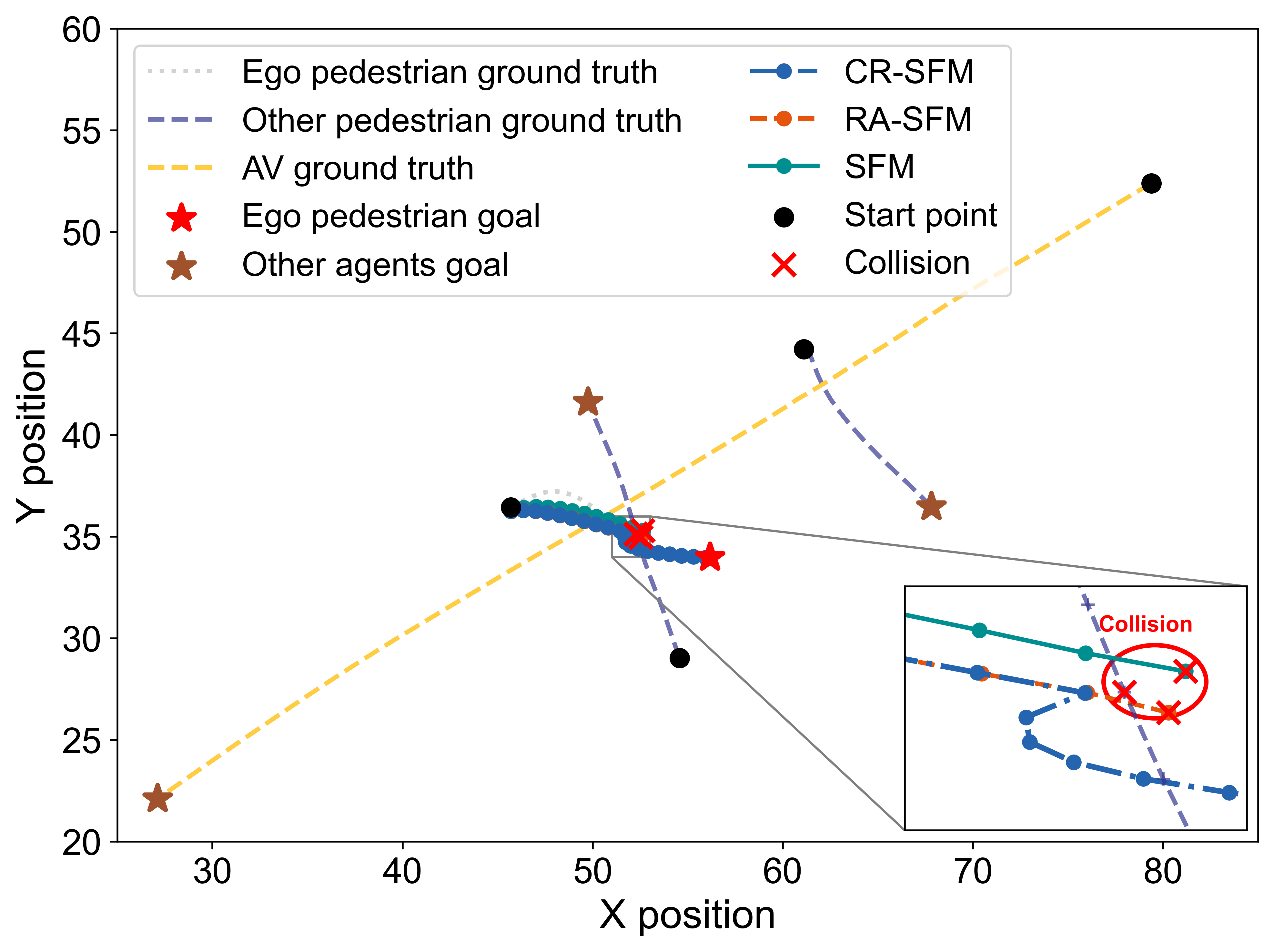}
        \caption{}
        \label{ped3}
    \end{subfigure}
    \caption{Qualitative examples of pedestrian trajectory simulation under different interaction scenarios. (a) A normal case in which all pedestrian trajectories are simulated by the proposed CR-SFM. (b) A comparison of three trajectories generated by the SFM, RA-SFM, and CR-SFM. The trajectory generated by the SFM results in a collision, while the RA-SFM and CR-SFM avoid the nearby agent. (c) The SFM and RA-SFM fail to trigger avoidance behavior, whereas the CR-SFM proactively detours before reaching the goal.}
    \label{pedtraj}
\end{figure}

To demonstrate how the proposed pedestrian model performs under different interaction conditions, three representative examples are presented in Figure \ref{pedtraj}. \par
Fig. \ref{ped1} illustrates a normal case in which the AV follows its ground-truth trajectory, while pedestrians are simulated using the proposed CR-SFM. As shown in the figure, all pedestrians safely reach their respective destinations without any collisions. The simulated trajectories (blue triangles) show high similarity to the ground-truth pedestrian trajectories (purple crosses), with only minor deviations. This confirms that the model’s ability to generate safe and goal-directed motion in interactive scenarios.\par
Fig. \ref{ped2} provides a scenario involving an explicit collision risk between the simulated pedestrian and a nearby agent. In this case, the AV and two pedestrians follow their ground-truth trajectories, while the third pedestrian’s trajectories are simulated by the SFM, RA-SFM, and CR-SFM. The zoomed-in subfigure reveals that the SFM-generated trajectory leads to a collision with another pedestrian, with collision point marked by a red cross. In contrast, both the RA-SFM and CR-SFM successfully avoid the hazard by slightly deviating the pedestrian’s path away from the surrounding agent, toward the right-hand side of the scene. This behavioral divergence highlights the benefit of adaptive social force modulation in RA-SFM and CR-SFM, enabling the agent to respond effectively to immediate physical threats. Notably, the trajectory generated by the CR-SFM displays a more pronounced shift toward the safer region, indicating a higher sensitivity to dynamic interaction contexts.  \par
Fig. \ref{ped3} further investigates the model’s behavior under ambiguous or latent risk conditions. Although no immediate physical danger is apparent, both the SFM and RA-SFM trajectories continue directly toward the goal and result in collisions, as shown in the zoomed-in area. In contrast, the CR-SFM trajectory shows a markedly different pattern: the pedestrian perceives elevated risk in the goal direction, detours toward a low-risk zone, and resumes goal-directed motion only after the uncertain agent has passed. \par
These behavioral differences stem from the models’ risk perception capabilities. The SFM relies on fixed interaction strengths based purely on distance, and thus cannot adapt to changing risk profiles. RA-SFM improves on this by adjusting repulsive forces according to observable physical cues such as proximity and velocity. However, when these cues remain below a critical threshold, the model fails to trigger sufficient repulsion, and the pedestrian proceeds along a collision-prone path without taking an evasive action.  \par
In contrast, the CR-SFM introduces cognitive uncertainty modeling, allowing the pedestrian to account not only for physical risk but also for uncertainty in the predicted behaviors of nearby agents. By quantifying the discrepancy between the expected and observed motion, CR-SFM captures the subjective uncertainty and then integrates it into the overall risk assessment. This allows the model to detect latent threats that are not evident through physical cues alone. Consequently, even when the physical risk is low, the elevated uncertainty can amplify the perceived danger and increase the total risk. When this combined risk becomes significant, the pedestrian proactively adjusts its trajectory, even temporarily moving away from the destination to avoid potential collisions. Once the perceived risk diminishes, the pedestrian returns to the original path and safely reaches the goal. This behavior illustrates the unique advantage of cognitive uncertainty modeling in CR-SFM. By integrating internal uncertainty estimation with external physical observations, the model is able to anticipate possible conflicts and make more informed decisions. As a result, CR-SFM generates more flexible, cautious, and human-like behavior in complex multi-agent environments, significantly enhancing both trajectory safety and realism. \par
In conclusion, the CR-SFM demonstrates strong performance for realistic pedestrian simulation, enabling reliable behavior modeling under diverse and challenging conditions.

\subsubsection{Quantitative Analysis}
Table \ref{pedresult} summarizes the quantitative performance of various pedestrian models in terms of ADE, FDE, and CR. The CV model yields the poorest performance, with an ADE of 1.5262 and an FDE of 2.8738, indicating large deviations from the ground-truth trajectories. It also records the highest collision rate, which is expected given that the CV model neglects goals, interactions, or environmental dynamics. In contrast, the SFM introduces interactive forces and goal attraction, significantly lowering the ADE and FED to 0.8310 and 0.6151, respectively. However, it still suffers from a non-negligible collision rate (3.7\%), highlighting its limitation in complex interactions due to its reliance on fixed interaction parameters. \par
\begin{table}[htbp]
\centering
\small
\caption{Statistic results of our pedestrian model compared with other approaches}
\label{pedresult}
\begin{tabularx}{\linewidth}{lXXX}
\toprule
Pedestrian Model & ADE & FDE & CR \\
\midrule
CV \cite{r1}                        & 1.5262       & 2.8738       & 0.0926      \\
SFM \cite{r41}                     & 0.8310       & 0.6151       & 0.0370      \\
RA-SFM                    & 0.8139       & 0.5247       & 0.0370      \\
\textbf{CR-SFM (Ours)}    & \textbf{0.7842} & \textbf{0.4416} & \textbf{0.0} \\
\bottomrule
\end{tabularx}
\end{table}
The RA-SFM enhances the traditional SFM by making the repulsive force adaptive to kinematic features such as distance, velocity and direction. This results in a slight improvement in the simulation accuracy, but the collision rate remains unchanged. Although RA-SFM allows for more adaptive motion generation, it still lacks the capacity to account for uncertainty in agent behaviors.By comparison, the proposed CR-SFM outperforms all baselines across all metrics. It obtains the lowest ADE (0.7842) and FDE (0.4416), reflecting its superior trajectory simulation accuracy. Notably, it achieves a zero collision rate, successfully avoiding all collisions in the evaluation scenarios. This performance is primarily attributed to the integration of cognitive risk modeling, which enables the pedestrian to not only react to visible physical cues but also anticipate potential conflicts arising from uncertainty in others’ behaviors. \par
In summary, the quantitative results demonstrate that the CR-SFM achieves reliable performance in both accuracy and safety. Compared to other methods, it reduces displacement errors while enhancing collision avoidance. These findings indicate that combining cognitive risk reasoning with adaptive social forces enables more flexible and robust generation of pedestrian behavior in dynamic and interactive environments.\par

\subsection{AV Decision-Making Model}
\subsubsection{Qualitative Analysis}
To clearly demonstrate how our AV decision-making model operates across different scenarios, two representative cases are presented below. \par
\begin{figure*}[t]
    \centering
    \begin{subfigure}[b]{0.49\linewidth}
        \centering
        \includegraphics[width=\linewidth]{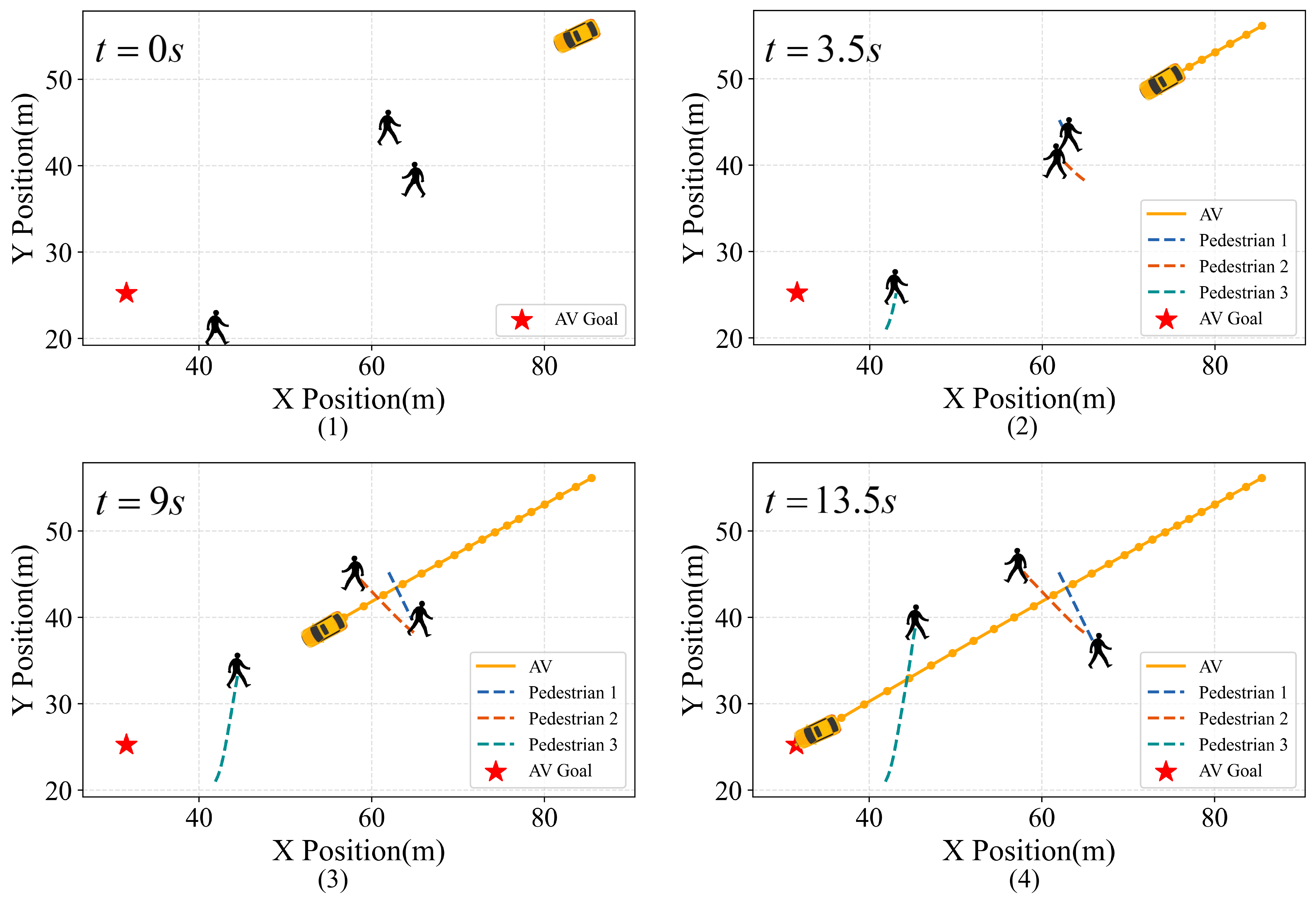}
        \caption{}
        \label{case1a}
    \end{subfigure}
    \hfill
    \begin{subfigure}[b]{0.49\linewidth}
        \centering
        \includegraphics[width=\linewidth]{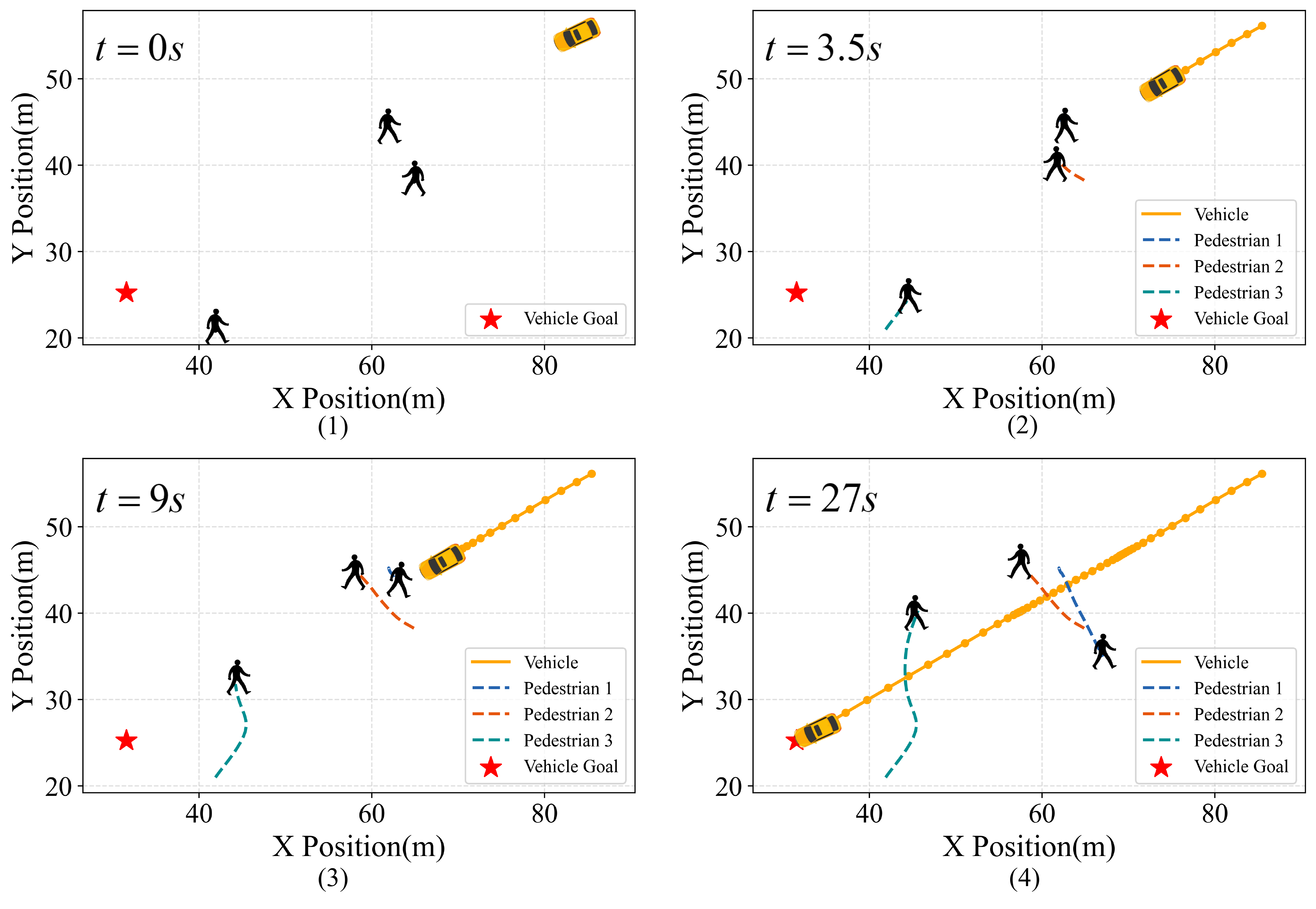}
        \caption{}
        \label{case1b}
    \end{subfigure}
    \vspace{1em}
    \begin{subfigure}[b]{0.49\linewidth}
        \centering
        \includegraphics[width=\linewidth]{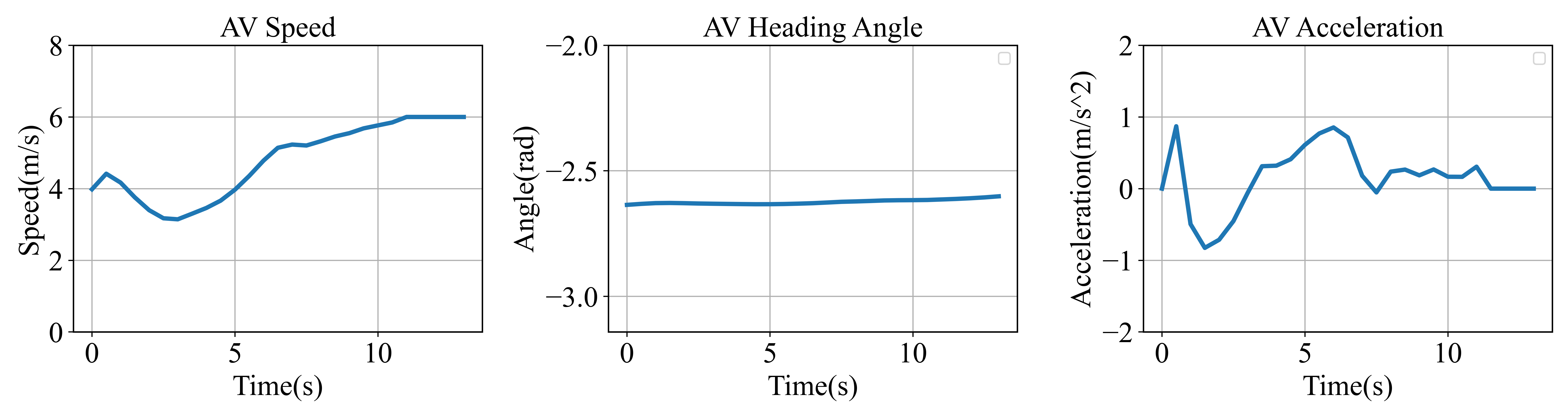}
        \caption{}
        \label{case1c}
    \end{subfigure}
    \hfill
    \begin{subfigure}[b]{0.49\linewidth}
        \centering
        \includegraphics[width=\linewidth]{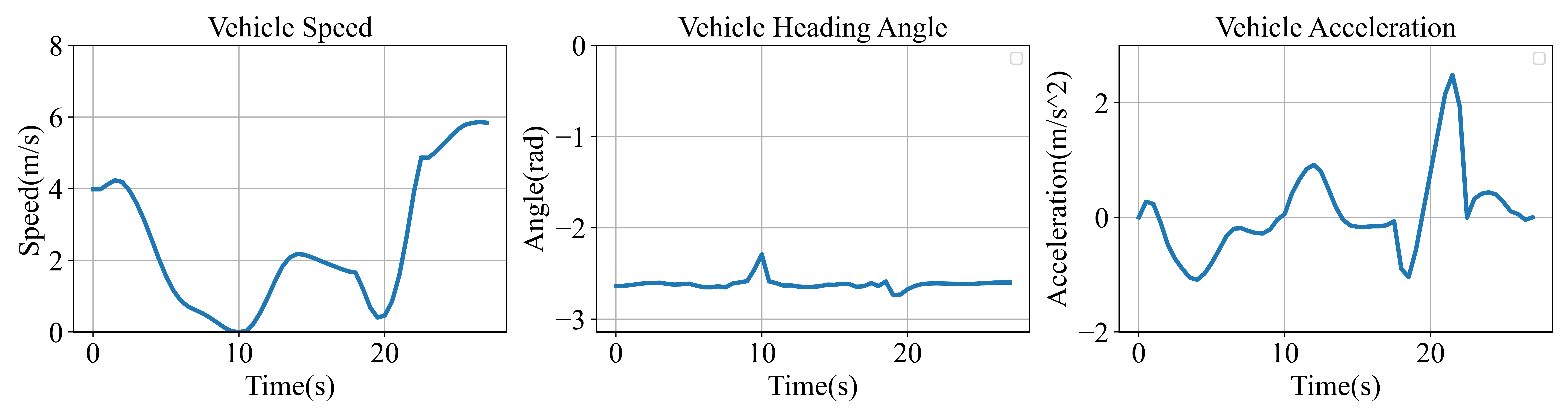}
        \caption{}
        \label{case1d}
    \end{subfigure}
    \caption{Comparison of the interaction process and dynamic state evolution between the proposed model and real driving data for Case 1. (a) Interaction simulation using the proposed model. (1) State at $t$ = 0 s: $v_{\text{AV}}$ = 3.981 m/s, $\theta$ = -2.635 rad, $a_{\text{AV}}$ = 0 m/s$^2$ (2) State at $t$ = 3.5 s: $v_{\text{AV}}$ = 3.144 m/s, $\theta$ = -2.630 rad, $a_{\text{AV}}$ = -0.057 m/s$^2$ (3) State at $t$ = 9 s: $v_{\text{AV}}$ = 5.455 m/s, $\theta$ = -2.619 rad, $a_{\text{AV}}$ = 0.265 m/s$^2$ (4) State at $t$ = 13.5 s: $v_{\text{AV}}$ = 6.0 m/s, $\theta$ = -2.601 rad, $a_{\text{AV}}$ = 0 m/s$^2$. (b) Interaction process captured from real driving data. (1) State at $t$ = 0 s: $v_{\text{veh}}$ = 3.981 m/s, $\theta$ = -2.635 rad, $a_{\text{veh}}$ = 0 m/s². (2) State at $t$ = 3.5 s: $v_{\text{veh}}$ = 3.578 m/s, $\theta$ = -2.601 rad, $a_{\text{veh}}$ = -0.903 m/s$^2$. (3) State at $t$ = 9 s: $v_{\text{veh}}$ = 0.266 m/s, $\theta$ = -2.597 rad, $a_{\text{veh}}$ = -0.281 m/s$^2$. (4) State at $t$ = 27 s: $v_{\text{veh}}$ = 5.838 m/s, $\theta$ = -2.599 rad, $a_{\text{veh}}$ = 0 m/s$^2$. (c) Evolution of AV states predicted by the proposed model. (d) Evolution of vehicle states recorded from real driving data.}
    \label{case_results1}
\end{figure*}

\textit{Case 1} provides an example where both the AV and the human driver decelerate early to avoid collisions with nearby pedestrians, as shown in Fig. \ref{case_results1}. At the beginning of the episode (Fig. \ref{case1a} and \ref{case1b}, top-left), the AV is at its initial position with three pedestrians distributed across the scene. Pedestrian 1 stands particularly close to the AV’s planned path, indicating a higher potential collision risk. In the simulation (Fig. \ref{case1a}), the AV gradually decelerates in response to the pedestrians’ anticipated behavior, slowing sufficiently to safely pass behind Pedestrians 1 and 2 before smoothly accelerating again once the path is clear. During the interaction with Pedestrian 3 (Fig. \ref{case1a}, bottom-left), the AV maintains a low, steady acceleration without significant speed reduction, reflecting its assessment of minimal risk and resulting in a smooth and continuous motion strategy. Finally, the AV successfully navigates through all pedestrians and reaches its goal without abrupt stops or collisions. \par
The pedestrian trajectories in the simulation (Fig. \ref{case1a}) are generated by our previously proposed pedestrian CR-SFM model, whereas those in Fig. \ref{case1b} are recorded from actual pedestrian movements. In the real-world scenario, although Pedestrian 1 chooses to stop and yield, the human driver also decelerates to a full stop, waiting until the pedestrian has completely crossed before resuming movement. After passing Pedestrians 1 and 2, the driver again reduces speed significantly to yield when approaching Pedestrian 3. These conservative actions lead to marked fluctuations in the vehicle’s speed and acceleration profiles, as illustrated in Fig. \ref{case1d}. In contrast, our proposed model achieves smoother and more stable driving behavior by applying sufficient deceleration without unnecessary stops. \par
\begin{figure*}[t]
    \centering
    \begin{subfigure}[b]{0.49\linewidth}
        \centering
        \includegraphics[width=\linewidth]{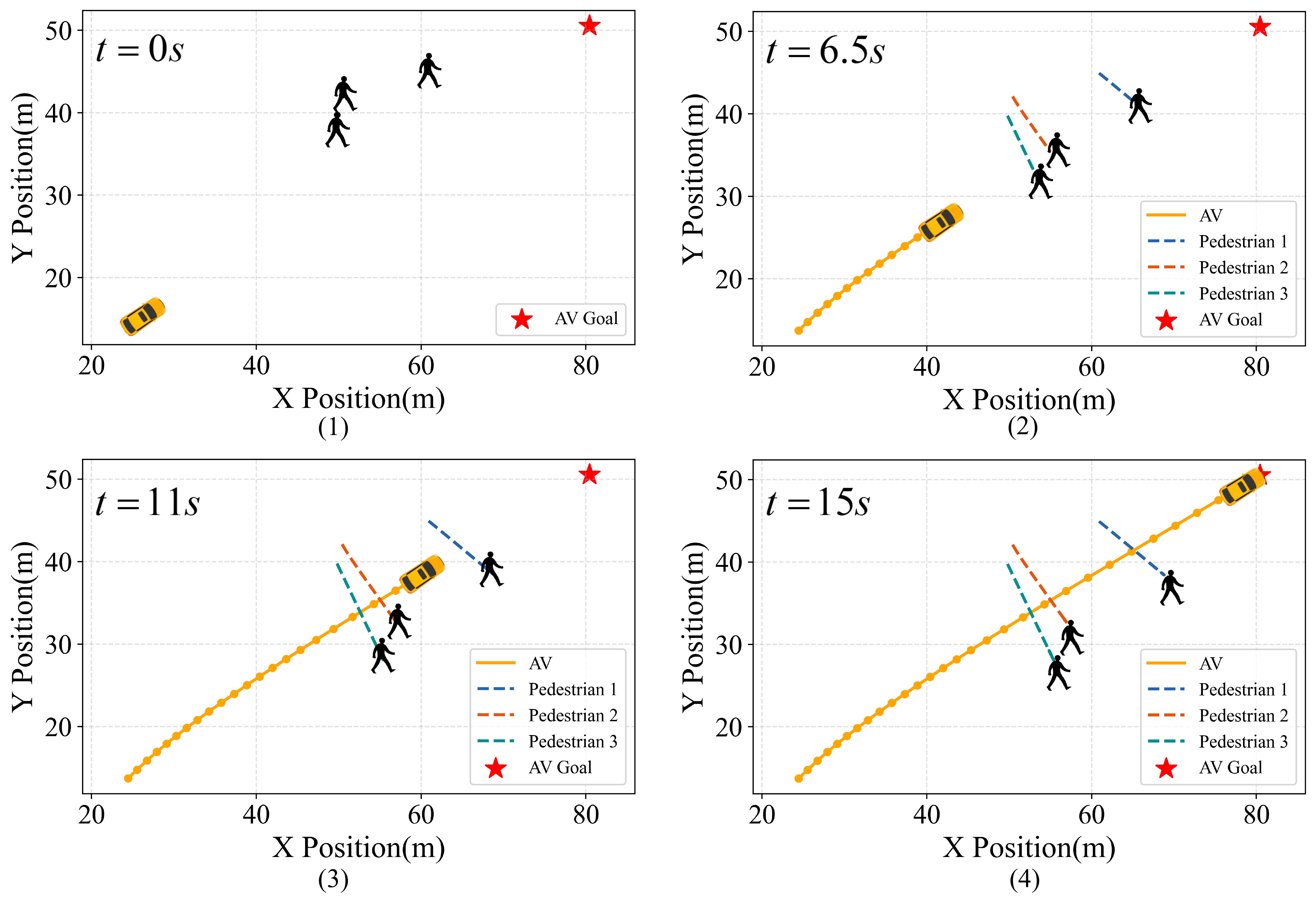}
        \caption{}
        \label{case2a}
    \end{subfigure}
    \hfill
    \begin{subfigure}[b]{0.49\linewidth}
        \centering
        \includegraphics[width=\linewidth]{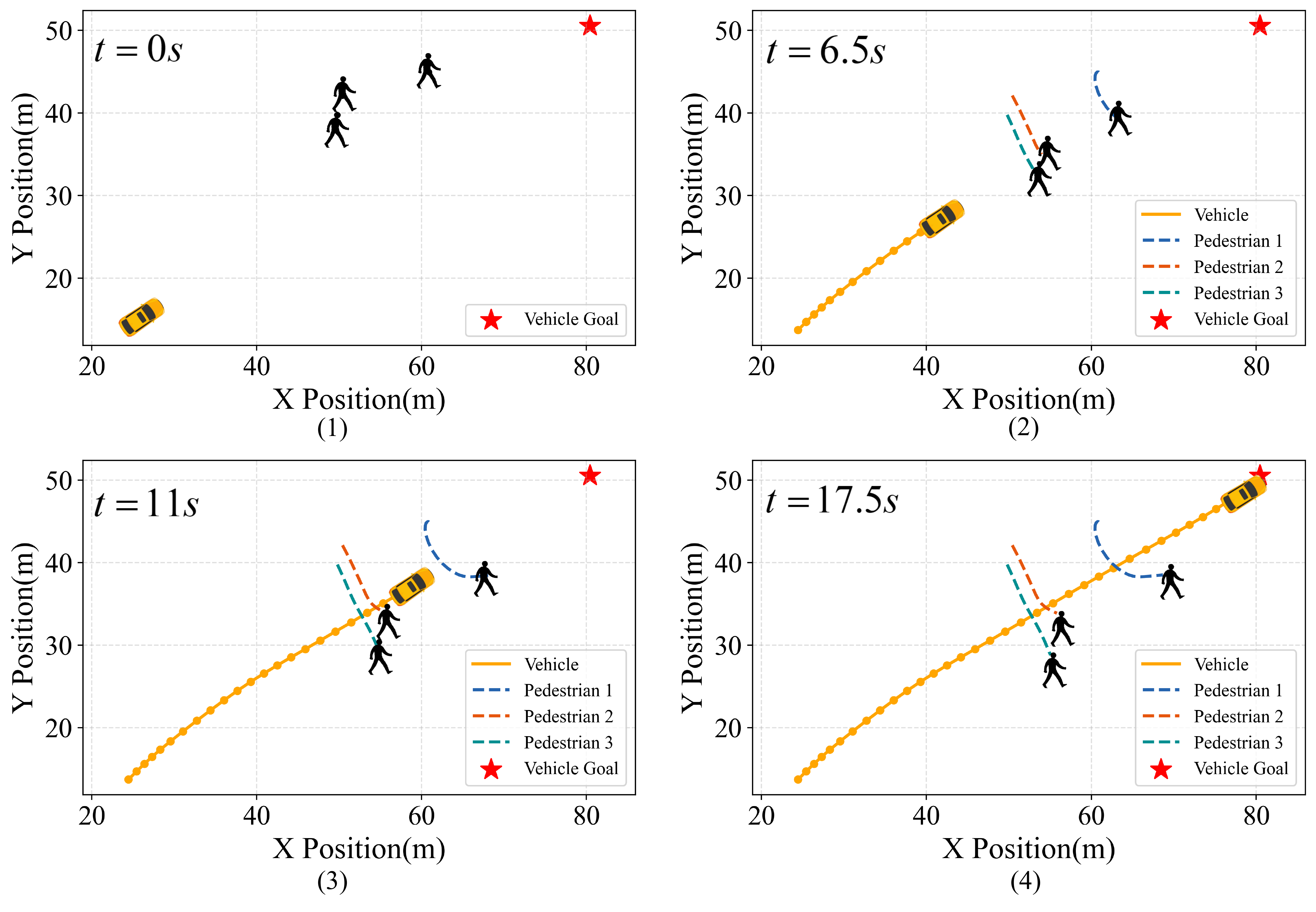}
        \caption{}
        \label{case2b}
    \end{subfigure}
    
    \vspace{1em}
    \begin{subfigure}[b]{0.49\linewidth}
        \centering
        \includegraphics[width=\linewidth]{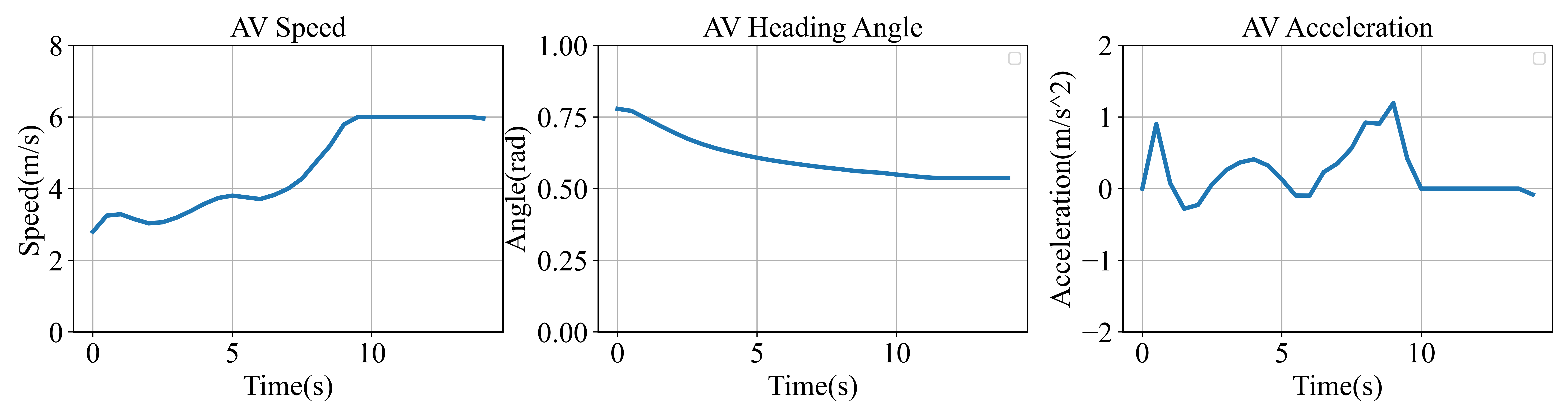}
        \caption{}
        \label{case2c}
    \end{subfigure}
    \hfill
    \begin{subfigure}[b]{0.49\linewidth}
        \centering
        \includegraphics[width=\linewidth]{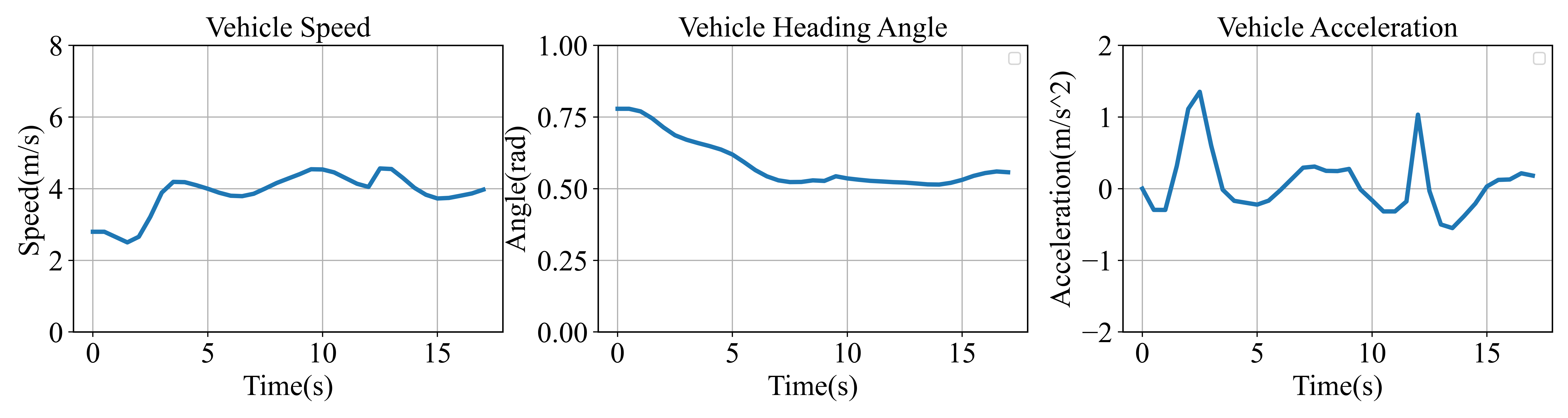}
        \caption{}
        \label{case2d}
    \end{subfigure} 
    \caption{Comparison of the interaction process and dynamic state evolution between the proposed model and real driving data for Case 2. (a) Interaction simulation using the proposed model. (1) State at $t$ = 0 s: $v_{\text{AV}}$ = 2.796 m/s, $\theta$ = 0.778 rad, $a_{\text{AV}}$ = 0 m/s$^2$ (2) State at $t$ = 6.5 s: $v_{\text{AV}}$ = 3.708 m/s, $\theta$ = 0.591 rad, $a_{\text{AV}}$ = -0.097 m/s$^2$ (3) State at $t$ = 11 s: $v_{\text{AV}}$ = 6.0 m/s, $\theta$ = 0.544 rad, $a_{\text{AV}}$ = 0 m/s$^2$ (4) State at $t$ = 15 s: $v_{\text{AV}}$ = 5.958 m/s, $\theta$ = 0.536 rad, $a_{\text{AV}}$ = -0.083 m/s$^2$. (b) Interaction process captured from real driving data. (1) State at $t$ = 0 s: $v_{\text{veh}}$ = 2.796 m/s, $\theta$ = 0.778 rad, $a_{\text{veh}}$ = 0 m/s$^2$. (2) State at $t$ = 6.5 s: $v_{\text{veh}}$ = 3.802 m/s, $\theta$ = 0.564 rad, $a_{\text{veh}}$ = -0.024 m/s$^2$. (3) State at $t$ = 11 s: $v_{\text{veh}}$ = 4.454 m/s, $\theta$ = 0.531 rad, $a_{\text{veh}}$ = -0.317 m/s$^2$. (4) State at $t$ = 17.5 s: $v_{\text{veh}}$ = 3.972 m/s, $\theta$ = 0.557 rad, $a_{\text{veh}}$ = 0.181 m/s$^2$. (c) Evolution of AV states predicted by the proposed model. (d) Evolution of vehicle states recorded from real driving data.}
    \label{case_results2}
\end{figure*}

\textit{Case 2} illustrates a situation in which both the AV and the driver maintain their speed to yield to pedestrians initially positioned at a greater distance, as shown in Fig. \ref{case_results2}. Early in this episode, both the AV and the human driver perceive that the pedestrians will enter their intended path. However, given the sufficient distance, they evaluate the situation as low-risk and opt to maintain a speed of approximately 3.5 m/s without significant braking. This behavior contrasts with Case 1, where the closer proximity of pedestrians required early deceleration to ensure safety. \par
Once the pedestrians have safely crossed and the potential risk is eliminated, the human driver continues at a steady speed of around 4 m/s, whereas the AV accelerates smoothly to its maximum velocity to efficiently reach its goal. Throughout the interaction, both the AV and the driver exhibit similar decision-making patterns, with consistent trends observed in their acceleration profiles and heading angle adjustments (Fig. \ref{case2c} and Fig. \ref{case2d}). These results suggest that the overall behavior of the proposed model closely mirrors that of the human driver, while showing smoother and more stable heading angle transitions.\par
In conclusion, the above results demonstrate that our proposed AV decision-making model can effectively select appropriate strategies based on the current scenario and state, adjusting its behavior accordingly to achieve safe yet efficient navigation during interactions with pedestrians.  \par

\subsubsection{Quantitative Analysis}
Table \ref{avtable} summarizes the quantitative performance of our proposed model (G-SAC-Cog) compared to the state-of-the-art method (UAW-PCG), two ablation variants (S-SAC and G-SAC-NoCog) and human driving data. Overall, our method achieves a success rate of 0.94, which is significantly higher than the comparative models (0.78–0.83). Moreover, it attains the lowest collision rate among the AV models at 0.06, indicating that our approach effectively enhances safety in pedestrian-rich environments. \par
\begin{table*}[]
\caption{Quantitative comparison results of our proposed AV model against other approaches and human driving data.}
\label{avtable}
\begin{tabular}{@{}ccccccc@{}}
\toprule
Model                    & Success  & Collision & Time out   & Average vehicle speed (m/s) & Average vehicle jerk (m/s$^3$) & \begin{tabular}[c]{@{}c@{}}Average maximum absolute \\ acceleration/deceleration (m/s$^2$)\end{tabular} \\ \midrule
UAW-PCG \cite{r2}                 & 0.83          & 0.17           & 0          & 4.137                      & 1.249                & 3.737                                                                                               \\
G-SAC-NoCog              & 0.83          & 0.11           & 0.06       & 4.035                      & 0.544                & 1.664                                                                                               \\
S-SAC                    & 0.78          & 0.22           & 0          & 4.312                      & 0.448                & 1.560                                                                                               \\
\textbf{G-SAC-Cog(Ours)} & \textbf{0.94} & \textbf{0.06}  & \textbf{0} & \textbf{4.470}             & \textbf{0.524}       & \textbf{1.621}                                                                                      \\
Human driver             & 1             & 0              & 0          & 2.924                      & 0.609                & 4.240                                                                                               \\ \bottomrule
\end{tabular}
\end{table*}
In terms of driving efficiency, our model achieves an average speed of 4.470 m/s, outperforming all comparative models and greatly exceeding that of the human driver (2.924 m/s). This superior efficiency can be attributed to the reinforcement learning framework applied in training AV models, which, under the given reward structure, incentivizes learned policies to maximize accumulated reward. Consequently, the AV maintains high speeds whenever no immediate collision risk is detected, enabling it to achieve higher efficiency without compromising safety, as evidenced by its low collision rate. \par
Regarding driving comfort, S-SAC attains the smoothest performance with the minimal average vehicle jerk (0.448 m/s$^3$). However, this comes with a high collision rate (0.22), suggesting that S-SAC tends to maintain speed even in close proximity to pedestrians, resulting in smoother but riskier behavior. Conversely, UAW-PCG shows the most erratic driving style, with an average vehicle jerk of 1.249 m/s$^3$. This instability likely stems from directly controlling vehicle speed, which forces the vehicle to make sudden and large acceleration or deceleration adjustments to match target speeds, resulting in frequent and abrupt speed variations. Our proposed model, which employs acceleration as the control action, achieves a well-balanced performance. Although its average jerk (0.524 m/s$^3$) is not the lowest, it reflects necessary adjustments for effective pedestrian avoidance and remains lower than that of the human driver (0.609 m/s$^3$), successfully balancing safety and comfort. When considering maximum absolute acceleration or deceleration, all AV models display substantially lower peaks compared to the human driver (4.240 m/s$^2$), indicating smoother control policies acquired through reinforcement learning. Notably, UAW-PCG shows a significantly elevated average maximum acceleration (3.737 m/s$^2$), again reflecting instability from using velocity as the direct action variable.  \par
In conclusion, these findings demonstrate that our proposed model achieves a superior balance between efficiency, safety, and driving comfort compared to existing methods. \par

\section{Conclusion}
This paper presents an innovative strategy to address AV decision-making challenges in shared urban spaces with multiple pedestrians. A cognitive uncertainty modeling approach inspired by the Free Energy Principle is introduced to emulate human reasoning and quantify cognitive uncertainty during social interactions. This cognitive modeling is integrated into the pedestrian social force model, enabling adaptive adjustment of goal-directed and repulsive forces by combining cognitive uncertainty with physical risk, which leads to more realistic pedestrian trajectories. Additionally, the same cognitive process guides the AV’s decision-making, where the fused measure of cognitive uncertainty and physical risk defines a dynamic, risk-aware adjacency matrix for the GCN within the SAC architecture, allowing the AV to perceive complex social dynamics and make human-like decisions. \par
Simulation results demonstrate that the proposed strategy significantly improves AV performance compared to the existing method, achieving safer, more efficient, and smoother behavior during interactions with pedestrians. Although the model shows promising results, the current research is limited to scenarios with a fixed number of pedestrians, which may not fully capture the variability of real-world urban environments. In the future, we plan to extend our strategy to handle scenarios with dynamically changing numbers of pedestrians, enabling the AV to adapt its strategy in real time as pedestrians enter or leave the scene.

\bibliographystyle{IEEEtran}
\bibliography{ref}
\begin{IEEEbiography}[{\includegraphics[width=1in,height=1.25in,clip,keepaspectratio]{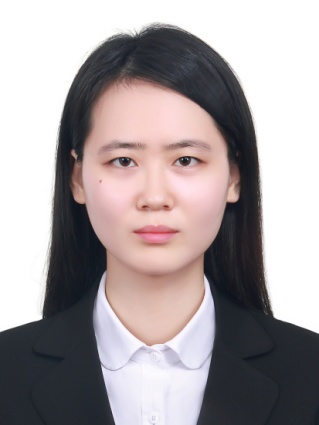}}]{Meiting Dang}received the B.S. and the M.S. degrees from Chang'an University, China, in 2017 and 2020, respectively. She is currently working toward the Ph.D. degree in James Watt School of Engineering with University of Glasgow, U.K. Her research interests include decision-making and planning of autonomous vehicles, autonomous vehicle-pedestrian interaction modeling based on game theory, and machine learning.
\end{IEEEbiography}

\begin{IEEEbiography}[{\includegraphics[width=1in,height=1.25in,clip,keepaspectratio]{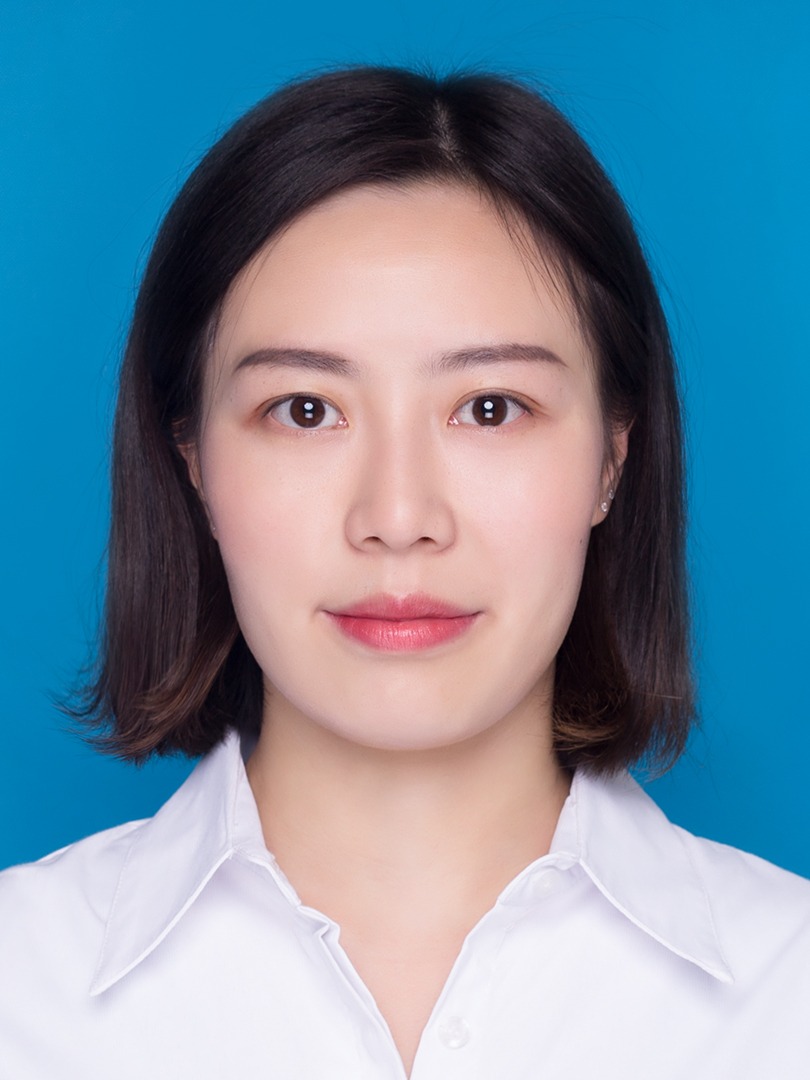}}]{Yanping Wu}received her B.S. degree from Southwest Jiaotong University, Chengdu, China, in 2017, and her M.S. degree in 2020. From 2020 to 2023, she worked as a Machine Learning Engineer at China Merchants Bank. She is currently pursuing a Ph.D. degree in Electronics and Electrical Engineering at the University of Glasgow. Her research interests include autonomous driving and artificial intelligence.
\end{IEEEbiography}

\begin{IEEEbiography}[{\includegraphics[width=1in,height=1.25in,clip,keepaspectratio]{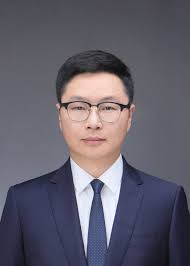}}]{Yafei Wang} (Member, IEEE) received the B.S. degree in internal combustion engine from Jilin University, Changchun, China, in 2005, the M.S. degree in vehicle engineering from Shanghai Jiao Tong University, Shanghai, China, in 2008, and the Ph.D. degree in electrical engineering from The University of Tokyo, Tokyo, Japan, in 2013. From 2008 to 2010, he was with automotive industry for nearly two years. From 2013 to 2016, he was a Postdoctoral Researcher with The University of Tokyo. He is currently a Professor of automotive engineering with the School of Mechanical Engineering, Shanghai Jiao Tong University. His research interests include state estimation and control for connected and automated vehicles.
\end{IEEEbiography}

\begin{IEEEbiography}[{\includegraphics[width=1in,height=1.25in,clip,keepaspectratio]{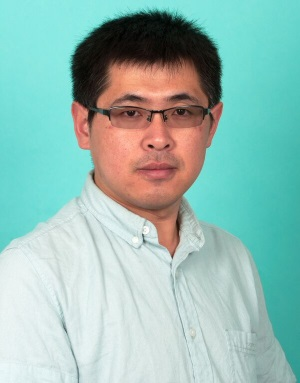}}]{Dezong Zhao} (Senior Member, IEEE) received the B.Eng. and M.S. degrees in control science and engineering from Shandong University, Jinan, China, in 2003 and 2006, respectively, and the Ph.D. degree in control science and engineering from Tsinghua University, Beijing, China, in 2010. He is currently a Senior Lecturer in autonomous systems with the School of Engineering,University of Glasgow, Glasgow, U.K. His research interests include connected and autonomous vehicles, machine learning, and control engineering. His work has been recognized by being awarded an EPSRC Innovation Fellowship and a Royal SocietyNewton Advanced Fellowship in 2018 and 2020, respectively.
\end{IEEEbiography}

\begin{IEEEbiography}[{\includegraphics[width=1in,height=1.25in,clip,keepaspectratio]{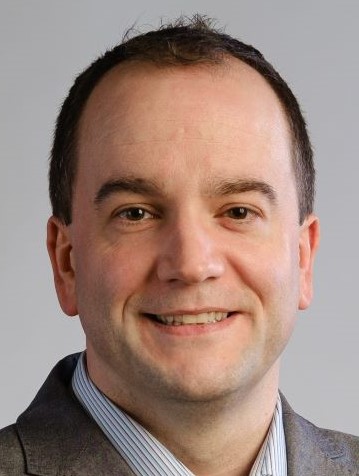}}]{David Flynn} (Senior Member, IEEE) is a Professor of Cyber Physical Systems (CPS) at the University of Glasgow, and Head of Research Division (HoRD) for Autonomous Systems and Connectivity within the James Watt School of Engineering. He has expertise in systems engineering, data analysis, digital twinning and digital technologies. He is an Honorary Professor of Heriot-Watt University, Fellow of the Royal Society of Edinburgh (FRSE), Fellow of the Institution of Engineers in Scotland (FIES,) and Fellow of the Royal Society for the Encouragement of Arts, Manufactures and Commerce (FRSA). He is the co-founder of the UKs EPSRC National Digital Twinning Hub (TransiT). His degrees include a BEng (Hons), 1st Class in Electrical and Electronic Engineering (2002), an MSc (Distinction) in Microsystems (2003) and a PhD in Microscale Magnetic Components (2007), from Heriot-Watt University, Edinburgh.
\end{IEEEbiography}

\begin{IEEEbiography}[{\includegraphics[width=1in,height=1.25in,clip,keepaspectratio]{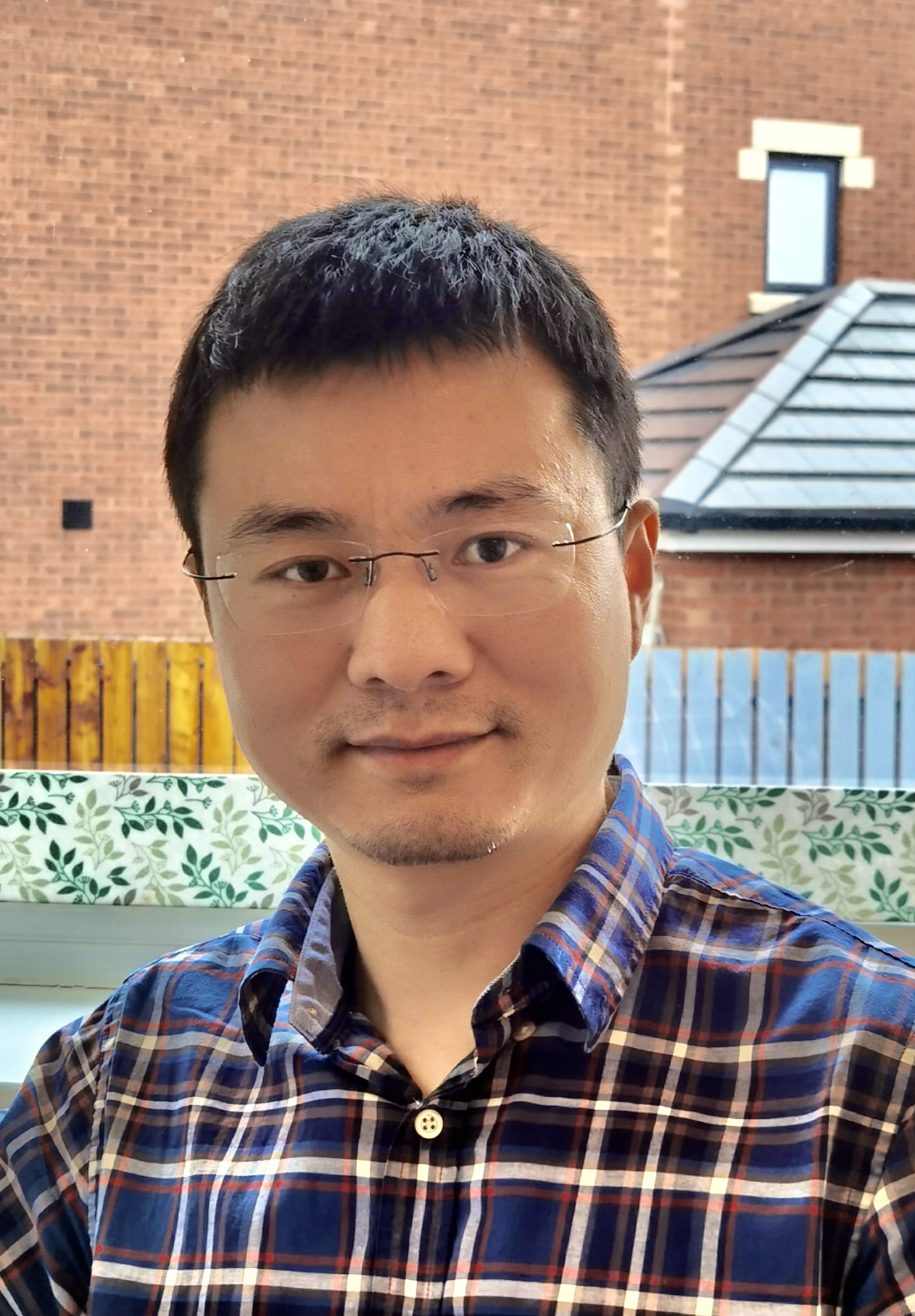}}]{Chongfeng Wei} (Senior Member, IEEE) received his Ph.D. degree in mechanical engineering from the University of Birmingham in 2015. He is now a Senior Lecturer (Associate Professor) at University of Glasgow, UK. His current research interests include decision-making and control of intelligent vehicles, human-centric autonomous driving, cooperative automation, and dynamics and control of mechanical systems. He is also serving as an Associate Editor of IEEE TITS, IEEE TIV, IEEE TVT, and Frontier on Robotics and AI.
\end{IEEEbiography}

\end{document}